\documentclass[11pt,twoside]{article}
\usepackage{fullpage}

\usepackage{epsf}
\usepackage{fancyhdr}
\usepackage{graphics}
\usepackage{graphicx}
\usepackage{psfrag}
\usepackage{microtype}
\usepackage{subfigure}
\usepackage{algorithmic}
\usepackage{color,xcolor}
\usepackage[linesnumbered,ruled]{algorithm2e}
\DontPrintSemicolon
\usepackage{color}

\usepackage{amsthm}
\usepackage{amsfonts}
\usepackage{amsmath}
\usepackage{amssymb,bbm}

\newcommand*\lrn[1]{\left\|#1\right\|}

\def\mI{\mathrm{I}}

\def\rd{\mathrm{d}}
\def\rT{{\rm T}}

\def\mI{\mathrm{I}}

\newcommand{\discretized}{\tilde}
\newcommand{\stepsize}{\eta}

\newcommand{\real}{\ensuremath{\mathbb{R}}}

\newcommand{\thetastar}{\ensuremath{\theta^*}}

\newcommand{\sphere}{\ensuremath{\mathbb{S}}}

\def\rT{{\rm T}}

\newcommand{\mydefn}{\ensuremath{:=}}

\newcommand{\defn}{:=}

\newcommand{\matsnorm}[2]{|\!|\!| #1 | \! | \!|_{{#2}}}
\newcommand{\vecnorm}[2]{\left\| #1\right\|_{#2}}

\newcommand{\inprod}[2]{\ensuremath{\langle #1 , \, #2 \rangle}}

\newcommand{\Exs}{\ensuremath{{\mathbb{E}}}}

\newtheoremstyle{named}{}{}{\itshape}{}{\bfseries}{.}{.5em}{\thmnote{#3's }#1}
\theoremstyle{named}

\theoremstyle{plain}

\newtheorem{theorem}{Theorem}
\newtheorem{proposition}{Proposition}

\newtheorem{lemma}{Lemma}

\newlength{\widebarargwidth}
\newlength{\widebarargheight}
\newlength{\widebarargdepth}

\makeatletter
\long\def\@makecaption#1#2{
        \vskip 0.8ex
        \setbox\@tempboxa\hbox{\small {\bf #1:} #2}
        \parindent 1.5em  
        \dimen0=\hsize
        \advance\dimen0 by -3em
        \ifdim \wd\@tempboxa >\dimen0
                \hbox to \hsize{
                        \parindent 0em
                        \hfil
                        \parbox{\dimen0}{\def\baselinestretch{0.96}\small
                                {\bf #1.} #2
                                }
                        \hfil}
        \else \hbox to \hsize{\hfil \box\@tempboxa \hfil}
        \fi
        }
\makeatother

\long\def\comment#1{}
\definecolor{battleshipgrey}{rgb}{0.52, 0.52, 0.51}
\definecolor{darkgray}{rgb}{0.66, 0.66, 0.66}
\definecolor{darkgreen}{rgb}{0.0, 0.2, 0.13}
\definecolor{darkspringgreen}{rgb}{0.09, 0.45, 0.27}
\definecolor{dukeblue}{rgb}{0.0, 0.0, 0.61}
\definecolor{olivedrab7}{rgb}{0.24, 0.2, 0.12}
\definecolor{darkblue}{rgb}{0.0, 0.0, 0.55}
\definecolor{darkscarlet}{rgb}{0.34, 0.01, 0.1}
\definecolor{candyapplered}{rgb}{1.0, 0.03, 0.0}
\definecolor{ao(english)}{rgb}{0.0, 0.5, 0.0}
\definecolor{applegreen}{rgb}{0.55, 0.71, 0.0}

\newcommand{\tens}{\ensuremath{\mbox{\tiny{tsr}}}}

\def\mD{{\rm D}}
\def\mQ{{\rm Q}}

\newcommand*{\pmu}{\boldsymbol{\mu}}

\newcommand{\E}{\mathbb E}

\newcommand*\Ep[1]{\mathbb{E}\left[#1\right]}
\usepackage[numbers]{natbib}
\usepackage{url}
\usepackage[colorlinks,linkcolor=magenta,citecolor=blue, pagebackref=true, backref=true]{hyperref}
 \renewcommand*{\backref}[1]{\ifx#1\relax \else Page #1 \fi}
\renewcommand*{\backrefalt}[4]{%
    \ifcase #1 \footnotesize{(Not cited.)}%
    \or        \footnotesize{(Cited on page~#2.)}%
    \else      \footnotesize{(Cited on pages~#2.)}%
    \fi}

\usepackage{nicefrac}
\usepackage{comment}

\usepackage{chngpage}

\usepackage{tabularx}%

\usepackage{enumitem}
\usepackage{booktabs}
\usepackage{caption}

\usepackage{bm,bbm}
\usepackage{mathtools}

\newcommand{\smooth}{L}

\newtheorem{assumption}{Assumption}
\newcommand{\Tmix}{T_{\mbox{\tiny{mix}}}}
\newcommand{\SUBFUNC}{\psi_{\theta, p}}

\newcommand{\discStep}{\ensuremath{\left(t-k\stepsize\right)}}

\makeatletter
\long\def\@makecaption#1#2{
        \vskip 0.8ex
        \setbox\@tempboxa\hbox{\small {\bf #1:} #2}
        \parindent 1.5em  
        \dimen0=\hsize
        \advance\dimen0 by -3em
        \ifdim \wd\@tempboxa >\dimen0
                \hbox to \hsize{
                        \parindent 0em
                        \hfil 
                        \parbox{\dimen0}{\def\baselinestretch{0.96}\small
                                {\bf #1.} #2
                                } 
                        \hfil}
        \else \hbox to \hsize{\hfil \box\@tempboxa \hfil}
        \fi
        }
\makeatother

\begin{document}

\begin{center}
{\bf{\LARGE{High-Order Langevin Diffusion Yields an Accelerated MCMC Algorithm}}}

\vspace*{.2in}
 {\large{
 \begin{tabular}{ccc}
  Wenlong Mou$^{\star, \diamond}$ & Yi-An Ma$^{\star, \diamond}$ &  Martin J. Wainwright$^{\diamond, \dagger, \ddagger}$\\
 \end{tabular}
 \begin{tabular}
 {cc}
  Peter L. Bartlett$^{\diamond, \dagger}$ & Michael I. Jordan$^{\diamond, \dagger}$
 \end{tabular}

}}

\vspace*{.2in}

 \begin{tabular}{c}
 Department of Electrical Engineering and Computer Sciences$^\diamond$\\
 Department of Statistics$^\dagger$ \\
 UC Berkeley\\
 \end{tabular}

 \vspace*{.1in}
 \begin{tabular}{c}
 Voleon Group$^\ddagger$
 \end{tabular}

\vspace*{.2in}

\today

\vspace*{.2in}
\begin{abstract}
  We propose a Markov chain Monte Carlo (MCMC) algorithm based on third-order Langevin dynamics
  for sampling from distributions with log-concave and smooth
  densities.  The higher-order dynamics allow for more flexible
  discretization schemes, and we develop a specific method that combines splitting with more accurate integration.  For a broad
  class of $d$-dimensional distributions arising from generalized
  linear models, we prove that the resulting third-order algorithm
  produces samples from a distribution that is at most
  $\varepsilon > 0$ in Wasserstein distance from the target distribution in
  $O\left(\frac{d^{1/4}}{ \varepsilon^{1/2}} \right)$ steps. This
  result requires only Lipschitz conditions on the gradient.  For
  general strongly convex potentials with $\alpha$-th order
  smoothness, we prove that the mixing time scales as $O \left(
  \frac{d^{1/4}}{\varepsilon^{1/2}} + \frac{d^{1/2}}{
    \varepsilon^{1/(\alpha - 1)}} \right)$.
\end{abstract}
\let\thefootnote\relax\footnotetext{$\star$ Wenlong Mou and Yi-An Ma
  contributed equally to this work.}
\end{center}


\section{Introduction}
\label{SecIntro}

Recent years have seen substantial progress in the theoretical
analysis of algorithms for large-scale statistical inference.  For
both the optimization algorithms used to compute frequentist point
estimates and the sampling algorithms that underpin Bayesian
inference, nonasymptotic rates of convergence have been obtained and,
increasingly, those rates include dimension dependence~\citep[see,
  e.g.,][]{Dalalyan_JRSSB,Moulines_ULA,
  Dalalyan_user_friendly,Xiang_underdamped,Xiang_overdamped,dwivedi2018log,MCMC_nonconvex,
  Variance_Reduction_theory}.  In particular, for the gradient-based
algorithms that have become the state-of-the-art in many large-scale
applications, the dimension dependence is generally linear or
sublinear, providing strong theoretical support for the deployment of
these algorithms in large-scale problems.

Although progress has been made in both optimization and sampling, the
latter has lagged the former, arguably because of the inherent
stochasticity of the sampling paradigm. Indeed, much of the recent
progress in both paradigms has involved taking a continuous-time point
of view, whereby algorithms are obtained as discretizations of
underlying continuous dynamical systems, and this line of attack is
more challenging for sampling methods.  For optimization algorithms
the continuous dynamics can be represented as ordinary differential
equations (ODEs)~\cite{Opt_Dyn,Weijie,Ashia,shi2018understanding},
whereas the underlying dynamics are characterized as stochastic
differential equations (SDEs) in the case of sampling
algorithms~\cite{Gareth,Dalalyan_JRSSB,Moulines_ULA}. The non-smooth
nature of the Brownian motion underlying these SDEs raises fundamental
challenges in carrying out the discretization that is needed to
transfer the continuous-time results to discrete time.

We focus on densities that can written in the form $p^*(\theta)
\propto \exp(-U(\theta))$, where the potential function $U: \real^d
\rightarrow \real$ is strongly convex and Lipschitz smooth.  There is
a substantial body of past work on sampling problems of this type;
among other results, it has been
shown~\cite{Xiang_underdamped,Xiang_Nonconvex,Dalalyan_underdamped,Yian_underdamped}
that a discretization of the second-order Langevin diffusion has
mixing time that scales as $O(\sqrt{d})$, which compares favorably to
the best known $O(d)$ scaling of the first-order Langevin
diffusion~\cite{Dalalyan_JRSSB,Moulines_ULA,
  Dalalyan_user_friendly}. The results were further improved by Shen
and Lee~\cite{shen2019randomized}, who used a uniform random time to
construct the low-bias estimator for an integral, leading to an
improved discretization scheme with $O (d^{1/3} / \varepsilon^{2/3})$
mixing time. Furthermore, Cao et al.~\cite{cao2020complexity} show for
second-order Langevin diffusions, this rate cannot be further improved
to achieve $\varepsilon$ discretization error.

However, if additional and relatively strong assumptions are imposed
on the density, then even faster rates of convergence can be
obtained~\cite{Mangoubi1,Mangoubi2,YinTat}.  Here we ask whether it is
possible to accelerate convergence of sampling algorithms beyond the
$O(d^{1/3})$ barrier in the general setting \emph{without} imposing
additional assumptions beyond strong convexity and Lipschitz smoothness. The
main contribution of our paper is an affirmative answer to this
question.

Let us provide some context for our line of attack and our
contributions.  As is well known from past
work~\cite{Dalalyan_JRSSB,Eberle_HMC}, the continuous-time Langevin
dynamics converge to the target distribution at an exponential rate,
with no dependence on dimension.  However, to be implemented with
digital computation, the continuous-time dynamics must be approximated
with a discrete-time scheme, leading to numerical error that does
scale with the dimension and the conditioning of the problem.  Direct
application of higher-order schemes to the Langevin diffusion is
hindered by the non-smoothness of the Brownian motion.  One way to
circumvent this problem is to augment the traditional Langevin
diffusion by moving to higher-order continuous dynamics.  In
particular, recent work has studied the the second-order (or
underdamped) Langevin algorithm, which lifts the original
$d$-dimensional space to a $2d$-dimensional space consisting of
vectors of the form $x=(\theta,r) \in \real^d \times \real^d$, and
considers a $2d$-dimensional collection of SDEs in these
variables~\cite{Xiang_underdamped,Xiang_Nonconvex,Dalalyan_underdamped,Yian_underdamped}.


There is a natural hierarchy of such lifted schemes, and this paper is
based on proposing and analyzing a carefully designed
\emph{third-order lifting}, to be described in
Section~\ref{SecThirdOrder}.  We provide a careful analysis of a
particular discretization of this third-order scheme, establishing
non-asymptotic bounds on mixing time for particular classes of
potential functions.

Our presentation begins with an analysis of potential functions that
have the following ridge-separable form:
\begin{align}
\label{eq:form-generalized-linear}  
U(\theta) = \sum_{i = 1}^n u_i \big(a_i^\rT \theta \big),
\end{align}
where $\{u_i\}_{i=1}^n$ are a collection of univariate functions, and
$\{a_i\}_{i=1}^n$ are a given collection of vectors in $\real^d$. Many
log-concave sampling problems that arise in statistics and machine
learning involve potential functions of this form. In particular,
posterior sampling in Bayesian generalized linear models, including
Bayesian logistic regression and one-layer neural networks, can be
written in the form~\eqref{eq:form-generalized-linear}. It is worth noticing that we do not impose any additional assumptions on the vectors $\{a_i\}_{i = 1}^n$. The ridge-separable form is needed only to make sure a one-dimensional integral to have close-form solution.

Given a distribution of the form~\eqref{eq:form-generalized-linear},
with $U$ strongly convex and smooth, we prove that $O(d^{\frac{1}{4}}
/ \varepsilon^{\frac{1}{2}})$ steps suffice to make the Wasserstein
distance between the sample and target distributions less than
$\varepsilon$. This is the first time that the $O(d^{1/3})$ barrier
for the log-concave sampling problem has been overcome without
additional structural assumptions on the data, even for the special case of
Bayesian logistic regression. The dependency on $\varepsilon$ is also
improved relative to the current state of the art.  It is worth
noticing that our analysis allows for arbitrary vectors
$\{a_i\}_{i=1}^n$ and functions $\{u_i \}_{i=1}^n$, as long as the
smoothness and strong convexity of $U$ are guaranteed. This is in
sharp contrast to previous work~\cite{Mangoubi1,Mangoubi2,YinTat} that
requires incoherence conditions on the data vectors $\{a_i\}_{i=1}^n$
and/or high-order smoothness conditions on the component functions
$\{u_i\}_{i=1}^n$.

We then tackle the more general setting in which the function $U$ need
not be ridge-separable as in
equation~\eqref{eq:form-generalized-linear}.  Assuming only that we
are given access to gradients from a black-box gradient oracle, we
show that the dimension dependency of our algorithm is $O(\sqrt{d})$,
but the dependency on final accuracy $\varepsilon$ associated with
this term can be adaptive to the smoothness assumptions satisfied by
$U$. In particular, we establish an upper bound on the mixing time of
$O \left( d^{\frac{1}{4}}/\varepsilon^{\frac{1}{2}} + d^{\frac{1}{2}}
/ \varepsilon^{\frac{1}{\alpha - 1}} \right)$, under $\alpha$-th order
smoothness of $U$. Thus, when the potential function $U$ satisfies
high-order smoothness conditions and a high-accuracy solution is
needed, this bound is favorable compared to existing $O(\sqrt{d} /
\varepsilon)$ rates.

The remainder of the paper is organized as follows.
Section~\ref{SecBackground} is devoted to background on the Langevin
diffusion, and various higher-order variants.  In
Section~\ref{SecMain}, we describe the third-order Langevin scheme
analyzed in this paper, and state our two main results:
Theorem~\ref{thm:finite-sum-main} for the special case of ridge-separable
functions, and Theorem~\ref{thm:abstract-highly-smooth} for general functions under black-box gradient access with additional smoothness.
Section~\ref{SecProofs} is devoted to the proofs of our main results,
with more technical aspects of the arguments deferred to the
appendices.  We conclude with a discussion in
Section~\ref{SecDiscussion}.


\section{Background and problem formulation}
\label{SecBackground}

In this section, we first introduce the class of sampling problems that are our focus, before turning to specific
sampling algorithms that are based on discretizations of diffusion processes.  We begin with the classical first-order discretization of Langevin diffusion, and then introduce the higher-order discretization that is the principal object of study of our work.


\subsection{A class of sampling problems}

We consider the problem of drawing samples from a distribution with
density written in the form $p^*(\theta) \propto \exp(- U(\theta))$.
The \emph{potential function} \mbox{$U: \real^d \rightarrow \real$} is
assumed to be strongly convex and smooth in the following sense:
\begin{assumption}[Strong convexity and smoothness]
  \label{AssAll}
The function $U$ is differentiable, and $m$-strongly convex and
$L$-smooth:
\begin{align}
\frac{m}{2} \||\theta' - \theta\|_2^2 \; \leq \; U(\theta') -
U(\theta) - \inprod{\nabla U(\theta)}{\theta' - \theta } \; \leq \;
\frac{L}{2} \|\theta' - \theta\|_2^2 \qquad \mbox{for all $\theta,
  \theta' \in \real^d$,}
  \end{align}
  where $m$ and $L$ are positive constants.
\end{assumption}
\noindent We say that the potential is \emph{$(m, L)$-convex-smooth}
when this sandwich relation holds.  The \emph{condition number} of the
problem is given by the ratio $\kappa \mydefn \frac{L}{m} \in [1,
  \infty)$.

Given an iterative algorithm that generates a random vector
$\theta^{(k)}$ at each step $k$, we use $\pi^{(k)}$ to denote the law
of $\theta^{(k)}$.  We are interested in the convergence $\pi^{(k)}$
to the measure $\pi^*$ defined by the target density $p^*$.  In order
to quantify closeness of the measures $\pi^{(k)}$ and $\pi^*$, we use
the Wasserstein-$2$ distance (see the
book~\cite{Villani_optimal_transport} for background). Given a pair of
distributions $p$ and $q$ on $\real^d$, a \emph{coupling} $\gamma$ is
a joint distribution over the product space $\real^d \times \real^d$
that has $p$ and $q$ as its marginal distributions.  We let $\Gamma(p,
q)$ denote the space of all possible couplings of $p$ and $q$.  With
this notation, the \emph{Wasserstein-$2$ distance} is given by
\begin{align}
  \label{EqnWasserstein}
\mathcal{W}_2^2 (p, q) & \defn \inf_{\gamma\in\Gamma(p,
  q)} \int_{\mathbb{R}^d\times\mathbb{R}^d} \lrn{x-y}_2^2 \rd
\gamma(x,y).
\end{align}
Given this definition, we obtain the notion of the $\varepsilon$-mixing time---it is
the minimum number of steps the algorithm takes to converge to within $\varepsilon$-close of the target measure $\pi^*$ in $\mathcal{W}_2^2$ distance:
\begin{align}
  \label{EqnMixingTime}
\Tmix(\varepsilon) \defn \min \left\{ k : \mathcal{W}_2 (\pi^{(k)},
\pi^*) \leq \varepsilon \right\}.
\end{align}


\subsection{First-order Langevin algorithm}

We refer to the stochastic process represented by the following
stochastic differential equation as \emph{continuous-time Langevin
  dynamics}:
\begin{align}
\label{EqnFirstOrderLangevin}
\rd \theta_t & = - \frac{1}{L} \nabla U(\theta_t) \ \rd t + \sqrt{2/L}
\ \rd B_t.
\end{align}
Here the reader should recall that $L$ is the smoothness parameter from
Assumption~\ref{AssAll}.

It is well known that continuous-time Langevin dynamics converges to
the target distribution $p^*$ exponentially quickly; moreover, under
Assumption~\ref{AssAll}, it is
known~\cite{Ledoux_log_Sobolev_diffusion} that the convergence rate is
independent of the dimension $d$.  However, after discretization using
the Euler scheme, the resulting algorithm---a discrete-time stochastic
process---has a mixing rate that scales as
$\mathcal{O}(d)$~\cite{Dalalyan_JRSSB}.  As this result makes clear,
the principal difficulty in high-dimensional sampling problems based
on Langevin diffusion is the numerical error that arises from the
integration of the continuous-time dynamics.  A classical response to
this problem is to introduce higher-order discretizations, but in the
setting of Langevin diffusion a major challenge arises---the
non-smoothness of the Brownian motion makes it difficult to control
high-order numerical errors.


\subsection{Underdamped (second-order) Langevin dynamics}

One way to proceed is to augment the dynamics to yield smoother
trajectories that are more readily discretized.  For example, the
\emph{second-order Langevin algorithm}, also known as the underdamped
Langevin algorithm, lifts the original $d$-dimensional space to a
\mbox{$2d$-dimensional} space consisting of vectors of the form \mbox{$x =
  (\theta,r) \in \real^d \times \real^d$,} and defines the following
\mbox{$2d$-dimensional} collection of SDEs:
\begin{align}
  \begin{cases}
    \rd \theta_t &= r_t \, \rd t \\
 \rd r_t &= - \frac{1}{L} \nabla U(\theta_t) \, \rd t - \xi r_t \,
 \rd t + \sqrt{2\xi/L} \; \rd B_t^r,
  \end{cases}
  \label{eq:underdamped_Langevin}
\end{align}
where $\xi > 0$ is an algorithmic parameter.

In the second-order Langevin dynamics determined by the
system~\eqref{eq:underdamped_Langevin}, the trajectory $\theta_t$ has
one additional order of smoothness compared to the Brownian motion
$B_t^r$.  As a result, it is possible to introduce higher-order
discretizations for the augmented dynamics. Examples of such
discretizations include Hamiltonian Monte
Carlo~\cite{NealHMC,Che+19_HMC} and underdamped Langevin
algorithms~\cite{Xiang_underdamped}, both of which are derived from
equation~\eqref{eq:underdamped_Langevin}.  These methods can provably
accelerate convergence; in particular, the underdamped Langevin
algorithm provides a convergence rate of $\mathcal{O}(\sqrt{d})$ when
the objective function $U$ is strongly convex and Lipschitz smooth.

\subsection{A third-order scheme}
\label{SecThirdOrder}

It is natural to ask whether one can further accelerate the
convergence of Langevin algorithms via higher-order dynamics, where we
expand the ambient space and drive the variable of interest via
higher-order integration of an SDE.  In order to pursue this question,
let us recall a generic recipe for constructing Markov dynamics with a
desired stationary distribution. Consider the family of SDEs of the
form
\begin{align}
\label{eq:generic}
\rd x_t = (\mD + \mQ) \nabla H(x_t) \ \rd t + \sqrt{2\mD} \ \rd B_t,
\end{align}
where $\mD$ is a constant positive semidefinite matrix, and $\mQ$ is a
constant skew-symmetric matrix.  It can be
shown~\cite{completesample,completeframework} that for any choice of
the matrices $(\mD, \mQ)$ respecting these constraints, the SDE in
equation~\eqref{eq:generic} has $p^*(x)\propto \exp(-H(x))$ as its
invariant distribution.

Within this general framework, note that the second-order Langevin
dynamics~\eqref{eq:underdamped_Langevin} are obtained by setting
$x_t=(\theta_t, r_t)$, and choosing
\begin{align*}
H(x_t) = H(\theta_t, r_t) = U(\theta_t) + \frac{L}{2} \lrn{r_t}^2,
\quad \mD = \frac{1}{L}
\begin{bmatrix}
0 & 0\\ 0 & \xi \mI
\end{bmatrix},
\quad \mbox{and} \quad
\mQ = \frac{1}{L}
\begin{bmatrix}
0 & \mI \\ -\mI & 0
\end{bmatrix}.
\end{align*}
Note that the positive semidefinite matrix $\mD$ has a zero top-left
block matrix (corresponding to the $\theta$ coordinates), which means
that $\theta_t$ is not directly coupled with the Brownian motion.

This observation motivates us to choose an even more singular $\mD$ matrix.  Beginning from the general equation~\eqref{eq:generic},
let $x_t = (\theta_t, p_t, r_t)$, and define the function
\mbox{$H(x_t) = U(\theta_t) + \frac{L}{2} \lrn{p_t}^2 + \frac{L}{2}
  \lrn{r_t}^2$,} along with the matrices
\begin{align*}
\mD = \frac{1}{L} \begin{bmatrix} 0 & 0 & 0\\ 0 & 0 & 0\\ 0 & 0 & \xi
  \mI
\end{bmatrix}, \quad \mbox{and}  \quad
\mQ = \frac{1}{L} \begin{bmatrix} 0 & \mI & 0\\ -\mI & 0 & \gamma
  \mI\\ 0 & - \gamma \mI & 0
\end{bmatrix}.
\end{align*}
Given these definitions, we set up a third-order form of Langevin dynamics as follows:
\begin{align}
  \label{eq:3order_dyn}
\begin{cases}
  \rd \theta_t = p_t \ \rd t \\
  \rd p_t = - \frac{1}{L} \nabla U(\theta_t) \ \rd t + \gamma r_t
  \ \rd t \\
  \rd r_t = - \gamma p_t \ \rd t - \xi r_t \ \rd t + \sqrt{2\xi/L}
  \ \rd B_t^r
\end{cases}.
\end{align}
The trajectory of $\theta_t$ under these third-order dynamics is
substantially smoother than the corresponding trajectory under the
underdamped Langevin dynamics; this higher degree of smoothness provides more control over discretization errors.  In particular, in our numerical analysis of
equation~\eqref{eq:3order_dyn}, we exploit the fact that the Brownian motion and $\nabla U$ are passed into the time derivative of two different variables, $r_t$ and $p_t$, respectively.  This allows us to adopt a splitting scheme that takes advantage of the structure of $U$ and
thereby provides an improvement in convergence rate relative to past work.  Indeed, in Section~\ref{SecMain} we prove that faster convergence is achieved with a proper choice of integration scheme.


\section{Main results}
\label{SecMain}

In this section, we describe our higher-order Langevin algorithm, and state two theorems that characterize its convergence rate.

\subsection{Discretized third-order algorithm}

We propose an algorithm, akin to the Langevin or underdamped Langevin
algorithm, that at every iteration generates a normal random variable
centered according to the previous iterate (see
Algorithm~\ref{alg:main}).  The algorithm is constructed via a
three-stage discretization scheme of the continuous-time Markov
dynamics~\eqref{eq:3order_dyn}. See Section~\ref{sec:disc} for a
detailed discussion of the discretization scheme.

Recalling the $(m,L)$-strong-convexity-smoothness condition given in
Assumption~\ref{AssAll}, we see that the potential function $U$ has a
unique global minimizer $\thetastar \in \real^d$ such that $\nabla
U(\thetastar) = 0$. We initialize our algorithm at a point $\theta_0$
satisfying $\vecnorm{\thetastar - \theta_0}{2} \leq \frac{1}{L}$. Such
a point can be found in $O(\sqrt{\kappa} \log (d \kappa) )$ gradient
evaluations using accelerated gradient methods~\cite{Nesterov_intro}.
Our algorithm generates a sequence of vector triples $x^{(k)} =
(\theta^{(k)}, p^{(k)}, r^{(k)})$ for $k = 1, 2, \ldots$ in a
recursive manner. Any instance of the algorithm is specified by a
stepsize parameter $\stepsize > 0$, two positive auxiliary parameters
$\gamma$ and $\xi$, and a function $\Delta U: \real^d \times \real^d
\rightarrow \real$.  We provide specific choices of these parameters
and the function $\Delta U$ in the theory to follow.
\begin{algorithm}
\caption{Discretized Third-Order Langevin Algorithm}\label{alg:main}
\begin{algorithmic}
\STATE Let $x^{(0)} = (\theta^{(0)}, p^{(0)}, r^{(0)}) = (\thetastar,
0, 0)$.  \FOR{$k=0,\cdots,N-1$} \STATE Sample ${x}^{(k+1)} \sim
\mathcal{N} \left( \pmu\left( {x}^{(k)} \right), \boldsymbol{\Sigma}
\right)$, where $\pmu$ and $\boldsymbol{\Sigma}$ are defined in
equation~\eqref{eq:mean_def} and \eqref{eq:cov_def}.  \ENDFOR
\end{algorithmic}
\end{algorithm}

Given the iterate $x^{(k)}$ at step $k$, the next iterate $x^{(k+1)}$ is obtained by drawing from a multivariate Gaussian distribution with mean $\pmu(x^{(k)})$, where
\begin{subequations}
\begin{align}
    \pmu\left({x}\right) \defn \left(
    \begin{array}{c}
         \theta - \frac{\stepsize}{2} \Delta U\left(\theta,p\right) +
         \mu_{12} p + \mu_{13} {r} \\ 
         - \Delta U\left(\theta,p\right) + \mu_{22} {p} + \mu_{23} {r} \\ 
         \mu_{31} \Delta U\left(\theta,p\right) + \mu_{32} {p} + \mu_{33} {r}
    \end{array}
    \right), \label{eq:mean_def}
\end{align}
and covariance
\begin{align}
    \boldsymbol{\Sigma} & \defn \frac{1}{L} \left(\begin{array}{ccc} 
    \sigma_{11} \cdot \mI_{d \times d} & \sigma_{12} \cdot \mI_{d
    \times d} & \sigma_{13} \cdot \mI_{d \times d} \\ 
    \sigma_{12} \cdot \mI_{d \times d} & \sigma_{22} \cdot \mI_{d \times d} 
    & \sigma_{23} \cdot \mI_{d \times d} \\ \sigma_{13} \cdot \mI_{d \times d} &
    \sigma_{23} \cdot \mI_{d \times d} & \sigma_{33} \cdot \mI_{d \times d}
    \end{array}\right). \label{eq:cov_def}
\end{align}
\end{subequations}
The constants $\mu_{12}$--$\mu_{33}$, as well as
$\sigma_{11}$--$\sigma_{33}$ above are entirely
determined by the triple $(\gamma, \xi, \stepsize)$; see
Appendix~\ref{appnd:discretization} for their explicit definitions.\\


\noindent We make a few remarks about the algorithm:
\begin{itemize}
  \item The vector $\Delta U (\theta, p)$ is chosen to be either an
    exact or approximate value of the integral \mbox{$\int_0^\stepsize
      \nabla U (\theta + t p) dt$.} As we discuss in the two versions
    of the main theorem, different choices of $\Delta U (\theta, p)$
    are available depending on the starting assumptions, and each such
    choice leads to a different mixing time bound.
  \item In each iteration, we only need to compute $\Delta
    U(\theta,p)$ once.  Below we provide choices of the function
    $\Delta U$ for which this step has equivalent computational cost
    with a gradient evaluation.
    \item When the stepsize $\stepsize$ is small, the leading terms in
      $(\mu (x^{(k)}), \Sigma)$ are the same as the dynamics in
      equation~\eqref{eq:3order_dyn}. However, the high-order
      correction terms are essential for achieving accelerated
      rates. This high-order scheme allows us to separate $\nabla U$,
      the only nonlinear part of the equation, and carry out a direct
      integration on a deterministic path.
   \item While our description allows for different choices of the
     parameters $\gamma$ and $\xi$, in our analysis, we adopt the
     choices $\gamma = \kappa$ and $\xi = 2 \kappa$.
\end{itemize}
%


\subsection{Guarantees for ridge-separable potentials}

We begin by describing and analyzing a version of our algorithm
applicable when the potential function $U$ is of ridge-separable
form~\eqref{eq:form-generalized-linear}.  In this case, the integral
$\int_0^\stepsize \nabla U (\theta + t p) dt$ can be computed exactly
using the Newton-Leibniz formula. This fact allows us to run
Algorithm~\ref{alg:main} with the choice
\begin{align}
\label{eq:deltau-ridge-separable}  
 \Delta U({\theta}, {p}) & \defn \frac{1}{L} \int_0^\stepsize \left(
 \sum_{i = 1}^n u_i' (a_i^\rT (\theta + t p)) a_i \right) dt \; = \;
 \frac{1}{L} \sum_i \left( u_i\left( a_i^\rT ( {\theta} + \stepsize
      {p} ) \right) - u_i\left( a_i^\rT {\theta} \right) \right)
      \frac{a_i}{a_i^\rT {p}}.
  \end{align}
We claim that an $O(d^{\frac{1}{3}})$ mixing time can be achieved in
this way.  More precisely, we have:
\begin{theorem}
\label{thm:finite-sum-main}
Let $U$ be an $(m, L)$-convex-smooth potential of the
form~\eqref{eq:form-generalized-linear}. Given a desired Wasserstein
accuracy $\varepsilon \in (0,1)$, suppose that we run
Algorithm~\ref{alg:main} with stepsize $\stepsize = c \kappa^{-
  \frac{11}{4}} d^{ - \frac{1}{4} } L^{ \frac{1}{4}} \varepsilon^{
  \frac{1}{2}}$, using the function $\Delta U (\theta, p)$ defined in
equation~\eqref{eq:deltau-ridge-separable}.  Then there is a universal
constant $C$ such that the mixing time is bounded as
\begin{align*}
    \Tmix(\varepsilon) \leq C \cdot \kappa^{\frac{19}{4}} (d /
    L)^{\frac{1}{4}} \big( \frac{1}{\sqrt{\varepsilon}} \big) \log
    \big(\frac{d \kappa}{\varepsilon} \big).
\end{align*}
\end{theorem}

Note that the result holds true for any potential function of the form equation~\eqref{eq:form-generalized-linear}, regardless of the distribution of the data points. In particular, we do not assume any form of incoherence assumptions as in~\cite{YinTat,Mangoubi1}; nor do we assume any conditions on the norm of vectors $a_i$. Furthermore, only the strong convexity and smoothness assumptions are used, without requiring high-order
smoothness assumptions. Many log-concave sampling problems of practical interest in statistical applications arise from a Gibbs measure defined by generalized linear potential functions. Under this setup, our result significantly improves the
previous best known $O(d^{1/3} / \varepsilon^{2/3})$
rate~\citep{shen2019randomized} in the dependency on both $\varepsilon$ and $d$. Finally, it is also worth noticing that the ridge-separable form is needed only to ensure the close-form expression~\eqref{eq:deltau-ridge-separable}. For a function that does not satisfy Eq~\eqref{eq:form-generalized-linear} but allows the close-form integration of $\Delta U (\theta, p)$, Theorem~\ref{thm:finite-sum-main} also applies.

As a caveat, we note that the stepsize used in running Algorithm~\ref{alg:main} depends on knowledge of $\kappa$ and $L$, which might not be unavailable in practice. An important direction for future work is to provide an automated mechanism for stepsize selection with similar guarantees.


\subsection{Guarantees under black-box gradient access}

We now turn to the more general setting, in which the potential function $U$ is no longer ridge
separable~\eqref{eq:form-generalized-linear}.  Suppose moreover that
we have access to $U$ only via a black-box gradient oracle, meaning
that we can compute the gradient $\nabla U(\theta)$ at any point
$\theta$ of our choice.  Under these assumptions, the closed form
integrator described in Algorithm~\ref{alg:main} is no longer
available.  However, by using Lagrange interpolation as an
approximation, we can still derive a practical high-order algorithm
that yields a faster mixing time.  In particular, while the mixing
time scales as $O(\sqrt{d})$, as with lower-order methods, we show
that the $\varepsilon$-dependency term can be adaptive to the degree of smoothness of the function $U$.

\paragraph{Lagrange interpolation:}

When the objective $U$ does not take the form of a generalized linear
function, we use Lagrange interpolating polynomials with Chebyshev
nodes~\cite{Numerical_Analysis} to approximate the function:
\begin{align*}
s \mapsto \nabla U\left( \frac{s - k \stepsize}{ \stepsize}
\hat{\theta}_{(k + 1) \stepsize} + \frac{(k + 1) \stepsize -
  s}{\stepsize} \discretized{\theta}_{(k)} \right) \quad \mbox{over
  the interval $s \in[k\eta,(k+1)\eta ]$.}
\end{align*}
In particular, for a given smoothness parameter $\alpha \in
\mathbb{N}_+$, the Chebyshev nodes are given by:
\begin{align*}
s_i = k\eta+\frac{\stepsize}{2} \left(1 + \cos \frac{2 i - 1}{2 \alpha}
\pi\right) \quad \mbox{for $i = 1, 2, \ldots, \alpha$.}
\end{align*}
The Chebyshev polynomial interpolation operator $\Phi$ takes as inputs
a scalar $t$, and a function $z: [0, \stepsize] \rightarrow \real^d$,
and returns the scalar $\Phi (t; z) \mydefn \sum_{i = 1}^\alpha
z(s_i) \prod_{j \neq i} \frac{t - s_i}{s_j - s_i}$.  Note that the
integral of this function over $t$ can be computed in closed form.

For each pair $(\theta, p)$ define the mapping $\SUBFUNC(s) = \nabla U
(\theta + s p)$ from $\real$ to $\real^d$.  Applying the interpolation
polynomial to this mapping, we define
\begin{align}
  \Delta U(\theta, p) & \defn \int_0^\stepsize \Phi (t; \SUBFUNC) dt.
\end{align}
Note that $t \mapsto \Phi (t; \SUBFUNC)$ is a polynomial function, and
hence the integral over $t$ can be computed exactly.  Computing this
integral requires $\alpha$ gradient evaluations in total, along with
$O(d)$ additional computational cost.  Thus, when the smoothness
$\alpha$ is viewed as a constant, the computational complexity is
order-equivalent to a gradient evaluation.
  
As is well known from the numerical analysis literature, higher-order
polynomial approximations are suitable to approximate functions that
satisfy higher-order smoothness condition.  In our analysis, we impose
a higher-order smoothness condition on $U$ in the following
way.  Note that the gradient $\nabla U(\theta)$ at any given $\theta$
is simply a vector, or equivalently a first-order tensor.  For a
first-order tensor $T$, its tensor norm is given by
$\matsnorm{T}{\tens}^{(1)} = \|T\|_2$, corresponding to the ordinary
Euclidean norm.  For a $k$-th order tensor $T$, we recursively define
its tensor norm as $\matsnorm{T}{\tens}^{(k)} \mydefn \sup_{v \in
  \sphere^{d - 1}} \matsnorm{T v}{\tens}^{(k - 1)}$, where
$\sphere^{d-1}$ denotes the Euclidean sphere in $d$-dimensions. With
this definition, the second-order tensor
$\matsnorm{\cdot}{\tens}^{(2)}$ norm for a matrix is equivalent to its
$\ell_2$-operator norm.

\begin{assumption}[High-order smoothness]
\label{assume:high-order-smooth}
For some $\alpha \geq 3$, the potential function $U$ is $\alpha$-th
order differentiable, and the associated tensor of derivatives
satisfies the bound
\begin{align*}
  \matsnorm{\nabla^{\alpha} U (x)}{\tens}^{(\alpha)} \leq
  L_\alpha^{\alpha - 1},
\end{align*}
for some $L_\alpha > 0$.
\end{assumption}
\noindent Note that in the special case $\alpha = 3$,
Assumption~\ref{assume:high-order-smooth} corresponds to a Lipschitz
condition on the Hessian function, as has been used in past analysis
of sampling algorithms. \\

\noindent In general, under a smoothness assumption of order $\alpha
\geq 2$, we have the following guarantee:
\begin{theorem}
\label{thm:abstract-highly-smooth}
Consider a potential $U$ satisfying Assumptions~\ref{AssAll} and
and~\ref{assume:high-order-smooth} for some $\alpha \geq 2$. Given a
desired Wasserstein accuracy $\varepsilon \in (0,1)$, suppose that we
run Algorithm~\ref{alg:main} with stepsize
\begin{subequations}
\begin{align}
\stepsize = c \cdot \min \left( \kappa^{- \frac{11}{4}} d^{ -
  \frac{1}{4} } L^{\frac{1}{4}} \varepsilon^{- \frac{1}{2}}, L_\alpha^{-1} L^{ \frac{1}{2}} \kappa^{-4} d^{-
  \frac{1}{2}} \varepsilon^{ \frac{1}{\alpha - 1}} \right),
\end{align}
using $\Delta U(\theta, p) = \int_0^\stepsize \Phi (t; \SUBFUNC) dt$
where $\SUBFUNC(s) = \nabla U(\theta + s p)$.  Then there is a
universal constant $C$ such that
\begin{align}
  \Tmix(\varepsilon) \leq C \cdot \max \left( L_\alpha \kappa^6
  \sqrt{d / L} \varepsilon^{- \frac{1}{\alpha - 1}}, \kappa^{\frac{19}{4}} (d /
  L)^{\frac{1}{4}} \varepsilon^{- \frac{1}{2}} \right).
\end{align}
\end{subequations}
\end{theorem}
\noindent We observe that the dimension dependence becomes $d^{1/2}$,
but the corresponding $\epsilon$ dependence is reduced to
$\epsilon^{-1/(\alpha-1)}$, where higher-order smoothness leads to
better accuracy dependence.


\subsection{Derivation of the discretization}
\label{sec:implementation}
\label{sec:disc}

In this section we provide a detailed derivation of Algorithm~\ref{alg:main} as a discretization scheme for the continuous-time
process~\eqref{eq:3order_dyn}.  Our overall approach involves a
combination of a splitting method and a high-order integration
scheme. More precisely, we reduce the problem of one-step simulation
of equation~\eqref{eq:3order_dyn} to (approximately) computing the
integral of $\nabla U$ along a straight line, using a three-stage
discretization scheme. Specifically, letting $\hat{g}_s$ be an
approximation for $\nabla U \left( \theta^{(k)} + \frac{s - k
  \stepsize}{\stepsize} p^{(k)} \right)$ and letting $\Delta U
(\theta^{(k)}, p^{(k)} ) \mydefn \int_{k \stepsize}^{(k + 1)
  \stepsize} \hat{g}_s ds$, the discretization error bound depends
on the accuracy with which $\hat{g}_s$ approximates $\nabla U$.
Depending on the assumptions imposed on the target distribution,
various choices of $\hat{g}$ can be used, the exact integration of
which leads to different choices of $\Delta U$.
Theorems~\ref{thm:finite-sum-main}
and~\ref{thm:abstract-highly-smooth} correspond to two instances of
this general approach.

Our three-stage discretization scheme is similar to classical splitting schemes for Langevin
dynamics~\cite{Splitting,Thermostat_Math}. We construct two
continuous-time processes: (a) an interpolation process
$(\discretized{\theta}, \discretized{p}, \discretized{r})$; and (b) an
auxiliary process $(\hat{\theta}, \hat{p}, \hat{r})$ defined over the
time interval $[k \stepsize, (k + 1) \stepsize)$.  At time $k
  \stepsize$, both processes are started from point $(\theta^{(k)},
  p^{(k)}, r^{(k)})$, and after one step of Algorithm~\ref{alg:main},
  we have $(\theta^{(k + 1)}, p^{(k + 1)}, r^{(k + 1)})
  =(\discretized{\theta}_{(k + 1) \stepsize}, \discretized{p}_{(k + 1)
    \stepsize}, \discretized{r}_{(k + 1) \stepsize})$.  It is worth noting that we only need to calculate the process $(\tilde{\theta}, \tilde{p}, \tilde{r})$ at time points that are integer multiples of $\stepsize$ in an algorithmic manner.

\paragraph{First stage:}  We begin
by constructing estimators $\hat{\theta}_{(k + 1)\stepsize}$
and $\hat{r}_{(k + 1) \stepsize}$ following the Ornstein-Uhlenbeck process:
\begin{align}
\label{eq:underdamp-euler}  
\begin{cases}
\rd \hat{\theta}_{t} = \discretized{p}_{k \stepsize} \rd t,\\ \rd
\hat{r}_t = - \gamma \discretized{p}_{k \stepsize} \rd t - \xi
\hat{r}_t \rd t + \sqrt{2 \xi / L} \rd B_t^r,
\end{cases} \quad  \mbox{for all  $t \in [k \stepsize, (k + 1) \stepsize]$.}
\end{align}

The values $(\hat{\theta}_{(k + 1) \stepsize}, \hat{r}_{(k + 1)
  \stepsize})$ are then used to calculate a high-accuracy result by
adding a correction term. We use the function $\hat{g}_s$ as an
approximation of the gradient
\begin{align*}
\nabla U( \hat{\theta}_s ) = \nabla U \left( \theta^{(k)} + (t - k \stepsize) \discretized{p}_{k \stepsize}  \right).
\end{align*}
It is worth noting that $\hat{g}_t$ approximates a function
along a fixed curve determined by $x_{k \stepsize}$, and has no
interaction with other variables nor the noise. This makes it possible
to obtain high-accuracy solutions to the equation by integration of a
deterministic and known function.

\paragraph{Second stage:}
In the second stage, we solve the system of differential equations
\begin{align}
  \label{eq:integration-p}
\begin{cases}
\rd \discretized{p}_t = - \frac{1}{L}\hat{g}_t \rd t + \gamma
\hat{r}_t dt,\\ \rd \hat{p}_t = - \frac{1}{L\stepsize}
\left(\displaystyle\int_{k \stepsize}^{(k + 1) \stepsize} \hat{g}_s
\rd s \right)\rd t + \gamma \hat{r}_t dt,
\end{cases}
\quad \forall t \in [k\stepsize, (k + 1) \stepsize].
\end{align}
Solving these equations amounts to integrating $\hat{g}_t$
and $\hat{r}_t$, whereas the quantity $\hat{p}_t$ corresponds to a
linear approximation of the integral component of $\tilde{p}_t$.  This
choice ensures that calculations for $\tilde{\theta}$ in our third
stage are straightforward. From equation~\eqref{eq:integration-p}, we
always have $\discretized{p}_{(k + 1) \stepsize} = \hat{p}_{(k + 1)
  \stepsize}$, which can be used as the value of $p^{(k + 1)}$ in
Algorithm~\ref{alg:main}.

\paragraph{Third stage:}  In the third stage, 
we use the estimator $\tilde{p}_t$ constructed from
equation~\eqref{eq:integration-p} in order to construct approximate
solutions to the following systems of SDEs:
\begin{align}
\label{eq:underdamp-correction-general}  
  \begin{cases}
    \rd \discretized{\theta}_{t} = \hat{p}_t \rd t\\ \rd
    \discretized{r}_t = - \gamma \hat{p}_t \rd t - \xi
    \discretized{r}_t \rd t + \sqrt{2 \xi/ L} \ \rd B_t^r,
    \end{cases} \quad t \in [k \stepsize, (k + 1) \stepsize].
\end{align}
Note that the Brownian motion $(B_t^r : t \geq 0)$ used in
process~\eqref{eq:underdamp-correction-general} must be the same as
that used in process~\eqref{eq:underdamp-euler}, so that the two
processes must be solved jointly. As shown in
Appendix~\ref{appnd:discretization}, we can carry out the integrations
in closed form, so as to obtain the explicit quantities required to
implement Algorithm~\ref{alg:main}.

\paragraph{Choices of approximator $\hat{g}_t$:} The three-stage discretization scheme that we have described is a
general framework, where we are free to choose different values of $\Delta (\theta, p) =
\int_0^\stepsize \hat{g}_{k \stepsize + s} \rd s$. The
choice of $\hat{g}_t$ is constrained by the need to make the integration exactly solvable,
and it has to serve as a good approximation for $\nabla U
(\hat{\theta}_t)$. If the potential function is of the form defined in
equation~\eqref{eq:form-generalized-linear}, we can simply take
$\hat{g}_t = \nabla U (\theta + t p)$, and the integration can be
carried out in closed form by the Newton-Leibniz formula.  Alternatively,
if $U$ is given by a black-box gradient oracle and satisfies
appropriate higher-order smoothness conditions, we can use the
Chebyshev node-interpolation method, and approximate $\nabla U (\theta
+ t p)$ using a polynomial in $t$. In such a case, the time integral
of $\hat{g}_t$ is also exactly solvable.


\section{Proofs}
\label{SecProofs}

In this section, we provide the proofs of our main results.  We begin
in Section~\ref{SecContinuousTime} by stating and proving a result
(Proposition~\ref{prop:contexp}) on the exponential convergence of the
third-order dynamics in continuous time.
Section~\ref{SecDiscreteTime} is devoted to our proofs of
Theorems~\ref{thm:finite-sum-main} and
Theorem~\ref{thm:abstract-highly-smooth} on the behavior of the
discrete-time algorithm.  In all cases, we defer the proofs of more
technical results to the appendices.


\subsection{Exponential convergence in continuous time}
\label{SecContinuousTime}

We begin by studying the process $\{x_t\}_{t \geq 0}$ defined by the
continuous-time third-order dynamics in
equation~\eqref{eq:3order_dyn}, with the particular goal of showing
convergence in the Wasserstein-$2$ distance.  In all of our
analysis---in this section as well as others---we make the choices
$\gamma=\kappa$ and $\xi=2\kappa$ in defining the dynamics.

It is known~\cite{completesample,completeframework} that the limiting
stationary distribution of the process $x_t = (\theta_t, p_t)$ has
a product form:
\begin{align*}
p^*(x) \propto e^{-H(x)} = e^{- U(\theta) - \frac{L}{2} \lrn{p}^2 -
  \frac{L}{2} \lrn{r}^2}.
\end{align*}
Our goal is to show that the distribution of $x_t = (\theta_t, p_t)$
converges at an exponential rate in the Wasserstein-$2$ distance, as
previously defined in equation~\eqref{EqnWasserstein}, to this
expanded target distribution.

In order to do so, we consider two processes following the third-order
dynamics~\eqref{eq:3order_dyn}, where the process $\{x_t\}_{t \geq 0}$
and $\{x^*_t\}_{t \geq 0}$ are started, respectively, from the initial
distributions $p_0$ and $p^*$. We then couple these two processes in a
synchronous coupling.  In order to establish a convergence rate, we make use of the following Lyapunov function:
\begin{subequations}
\begin{align}
  \label{EqnDefnLyapunov}
t \mapsto \mathcal{L}_t = \inf_{\zeta_t\in\Gamma\left(p_t, p^*\right)}
\E_{(x_t,x^*)\sim\zeta_t} \left[(x_t - x^*)^\rT S (x_t - x^*)\right],
\end{align}
where the symmetric matrix $S$ is given by
\begin{align}
  \label{eq:S_def}
  S & \mydefn
  \left(
\begin{array}{ccc}
 \frac{\kappa ^7+3 \kappa ^4+5 \kappa ^3+\kappa +1}{4 \kappa ^5} \mI &
 \frac{\kappa}{2} \mI & \frac{1}{4} \left(1-\frac{1}{\kappa
   ^3}-\frac{1}{\kappa ^4}\right) \kappa \mI \\ \frac{\kappa }{2} \mI
 & \frac{4 \kappa ^4+6 \kappa ^3+\kappa +1}{4 \kappa ^4} \mI &
 \frac{\kappa +1}{2 \kappa} \mI \\ \frac{1}{4} \left(1-\frac{1}{\kappa
   ^3}-\frac{1}{\kappa ^4}\right) \kappa \mI & \frac{\kappa +1}{2
   \kappa} \mI & \frac{\kappa +2}{4 \kappa} \mI \\
\end{array}
\right).
\end{align}
\end{subequations}

\noindent With this setup, our main result on the continuous-time
dynamics is the following:
\begin{proposition}
\label{prop:contexp}  
Let $x_0$ and $x^*_0$ follow the laws of $p_0$ and $p^*$,
respectively.  Then the process $\{x_t\}_{t \geq 0}$ defined by the
dynamics~\eqref{eq:3order_dyn} satisfies the bound
\begin{align*}
    \inf_{\zeta_t\in\Gamma\left(p_t, p^*\right)} \E_{(x_t,x^*)\sim\zeta_t} \left[(x_t - x^*)^\rT S (x_t -
      x^*)\right] \leq e^{- \frac{t}{5\kappa^2+50}}
    \inf_{\zeta_0\in\Gamma\left(p_0, p^*\right)} \E_{(x_0,x^*)\sim\zeta_0} \left[(x_0 - x^*)^\rT S (x_0 -
      x^*)\right].
\end{align*}
\end{proposition}
\noindent As shown in Lemma~\ref{lemma:eig_S}, to be stated in the
next section, the eigenvalues of $S$ lie in the interval
$[1/(5\kappa), \; \kappa^2 + 10 ]$.  Thus,
Proposition~\ref{prop:contexp} implies convergence in the
Wasserstein-$2$ distance at an exponential rate.

The remainder of this section is devoted to the proof of
Proposition~\ref{prop:contexp}.  The first step in the proof involves
establishing a differential inequality for the Lyapunov function
$\mathcal{L}_t$ via the coupling technique.
\begin{lemma}
\label{lemma:cont}
Let the processes $\{x_t\}$ and $\{x^*_t\}$ follow the third-order
dynamics~\eqref{eq:3order_dyn} with initial conditions $x_0$ and
$x^*_0 \in \mathbb{R}^{3d}$.  Then there exists a coupling
$\bar{\zeta}\in\Gamma\left(p_t(x_t|x_0), p^*_t(x^*_t|x^*_0)\right)$ of
the laws of $x_t$ and $x^*_t$ such that
\begin{align}
   \label{eq:contraction}  
    \frac{\rd}{\rd t} (x_t - x^*_t)^\rT S (x_t - x^*_t) \leq -
    \frac{1}{5\kappa^2+50} (x_t - x^*_t)^\rT S (x_t - x^*_t) \qquad
    \mbox{for all $(x_t, x^*_t) \sim \bar{\zeta}$.}
\end{align}
\end{lemma}
\noindent The proof of Lemma~\ref{lemma:cont}, given in
Section~\ref{SecProofLemmaCont}, is based on the synchronous coupling
technique, in which two processes are coupled based on the same
underlying Brownian motion.

Taking Lemma~\ref{lemma:cont} as given, we can now complete the proof
of Proposition~\ref{prop:contexp}.  Applying Gr\"onwall's lemma to
equation~\eqref{eq:contraction} yields
\begin{align*}
(x_t - x^*_t)^\rT S (x_t - x^*_t) \leq e^{-t/(5\kappa^2+50)}
  \left((x_0 - x^*_0)^\rT S (x_0 - x^*_0)\right).
\end{align*}
Noting that $\hat{\zeta}_t(x_t,x_t^*) = \E_{(x_0,x^*_0)\sim\zeta_0^*}
\left[ \bar{\zeta}(x_t, x^*_t | x_0, x_0^*) \right]$ is a coupling, we
find that
\begin{align*}
\lefteqn{\inf_{\zeta_t\in\Gamma\left(p_t, p^*_t\right)}
  \E_{(x_t,x^*_t)\sim\zeta_t} \left[(x_t - x^*_t)^\rT S (x_t -
    x^*_t)\right]} \\ &\leq \E_{(x_t,x^*)\sim\hat{\zeta}_t} \left[(x_t
  - x^*_t)^\rT S (x_t - x^*_t)\right] \\ &=
\E_{(x_0,x^*_0)\sim\zeta_0^*} \left[ \E_{(x_t,x^*)\sim\bar{\zeta}(x_t,
    x^*_t | x_0, x_0^*)} \left[(x_t - x^*_t)^\rT S (x_t -
    x^*_t)\right] \right] \\ &\leq \E_{(x_0,x^*_0)\sim\zeta_0^*}
\left[ e^{-t/(5\kappa^2+50)} \left((x_0 - x^*_0)^\rT S (x_0 -
  x^*_0)\right) \right] \\ &= e^{-t/(5\kappa^2+50)}
\inf_{\zeta_0\in\Gamma\left(p_0, p^*\right)} \E_{(x_0,x^*)\sim\zeta_0}
\left[(x_0 - x^*)^\rT S (x_0 - x^*)\right],
\end{align*}
which establishes the bound
in Proposition~\ref{prop:contexp}.

\subsubsection{Proof of Lemma~\ref{lemma:cont}}
\label{SecProofLemmaCont}

We prove Lemma~\ref{lemma:cont} by choosing a synchronous
coupling $\bar{\zeta}\in\Gamma\left(p_t(x_t|x_0),
p^*_t(x^*_t|x^*_0)\right)$ for the laws of $x_t$ and $x^*_t$.  (A
synchronous coupling simply means that we use the same Brownian motion
$B_t^r$ in defining both $x_t$ and $x^*_t$.)  We then obtain that for
any pair $(x_t,x^*_t)\sim\bar{\zeta}$,
\begin{align}
\left(
\begin{array}{c}
\rd (\theta_t - \theta^*_t) \\
\rd (p_t - p^*_t) \\
\rd (r_t - r^*_t) 
\end{array}
\right)
= (\mD + \mQ) 
\left(
\begin{array}{c}
\nabla U(\theta_t) - \nabla U(\theta^*_t) \\
L (p_t - p^*_t) \\
L (r_t - r^*_t)
\end{array}
\right)
\rd t .
\end{align}
Since $U$ is a Lipschitz-smooth function defined on $\mathbb{R}^d$, an open convex domain, we can use the mean-value theorem for vector-valued functions and write $\nabla U(\theta_t) -
\nabla U(\theta^*_t) = H_t (\theta_t - \theta^*_t)$, where
\begin{align}
  \label{EqnHmat}
\displaystyle H_t = \int_0^1 \nabla^2 U\left(
\theta^*_t + \lambda \left( \theta_t - \theta^*_t \right) \right)\rd\lambda.
\end{align}
We obtain that $\rd (x_t - x^*_t) = M_t (x_t - x^*_t) \rd
t$, where the matrix $M_t$ takes the form
\begin{align*}
M_t & \mydefn \left(
    \begin{array}{ccc}
        0 & \mI & 0\\
        -\frac{H_t}{L} \mI & 0 & \gamma \mI\\
        0 & - \gamma \mI & -\xi \mI
    \end{array}\right).
\end{align*}
Consequently, the derivative of the function $t \mapsto (x_t -
x^*_t)^\rT S (x_t - x^*_t)$ is given by
\begin{align*}
\frac{\rd}{\rd t} (x_t - x^*_t)^\rT S (x_t - x^*_t)
&= 2 (x_t-x^*_t)^\rT S M_t (x_t - x^*_t) \\
&= (x_t-x^*_t)^\rT (S M_t + M_t^\rT S) (x_t - x^*_t).
\end{align*}
In order to proceed, we need to relate the eigenvalues of the matrix
$S M_t + M_t^\rT S$ to those of $S$.  The following lemmas allow us to
carry out this conversion:
\begin{lemma}
\label{lemma:eig}
For any $\kappa=L/m\geq 1$ and matrix $H_t$ of the
form~\eqref{EqnHmat} such that $m \mI \preceq H_t \preceq L \mI$, the
eigenvalues of $S M_t + M_t^\rT S$ are smaller than $-1/5$.
\end{lemma}

\begin{lemma}
\label{lemma:eig_S}
For any $\kappa=L/m \geq 1$ and $m \mI \preceq H_t \preceq L \mI$, the
eigenvalues of the matrix $S$ lie in the interval $[\frac{1}{5\kappa},
  \; \kappa^2 + 10]$.
\end{lemma}

\noindent Using these two lemmas, it follows that for any pair of
random variables $(x_t, x^*_t) \sim \bar{\zeta}$, we have
\begin{align*}
 \frac{\rd}{\rd t} (x_t - x^*_t)^\rT S (x_t - x^*_t) & =
 (x_t-x^*_t)^\rT (S M_t + M_t^\rT S) (x_t - x^*_t) \nonumber\\ &
 \stackrel{(i)}{\leq} -\frac{1}{5} \lrn{x_t - x^*_t}^2 \nonumber \\
& \stackrel{(ii)}{\leq} - \frac{1}{5\kappa^2+50} (x_t - x^*_t)^\rT S
 (x_t - x^*_t), 
\end{align*}
where inequality (i) follows from Lemma~\ref{lemma:eig} and inequality
(ii) follows from the upper bound on the eigenvalues of $S$ in
Lemma~\ref{lemma:eig_S}.  This completes the proof of
Lemma~\ref{lemma:cont}. \\

\noindent Finally, we turn to the proofs of Lemmas~\ref{lemma:eig}
and~\ref{lemma:eig_S}.

  
\subsubsection{Proof of Lemma~\ref{lemma:eig}}

In order to simplify notation, we first define $\{l_k \ | \ k=1,\cdots
d\}$ as the eigenvalues of $H_t/L$.  Since $H_t$ is the Hessian of the
potential function $U$ (which is $m$-strongly convex and $L$-Lipschitz
smooth), its eigenvalues are always bounded above and below: $1/\kappa
\leq l_k \leq 1$.

Next we calculate the eigenvalues $\{\lambda_k\}$ of $SM_t+M_t^\rT S$: $\lambda_0=-1$ and
\[
\lambda_k^{\pm} = -\frac{2 (l_k+1/\kappa) \pm (l_k-1/\kappa ) \sqrt{
g_1(\kappa) } } {4/\kappa },
\]
where we define $g_1(\kappa)$ in Appendix~\ref{appnd:g} to simplify the notation.

We first upper bound $\lambda_k^{+}$.  Since $(l_k-1/\kappa ) \sqrt{
  g_1(\kappa) } \geq 0$, we find that
\begin{align*}
  \lambda_k^{+} \leq -\frac{2 (l_k+1/\kappa) } {4/\kappa } \leq -1.
\end{align*}
Turning to the bound on $\lambda_k^{-}$, we can rewrite it as:
\begin{align*}
\lambda_k^{-} = - \frac{\kappa}{4} \left( \left( 2 - \sqrt{
  g_1(\kappa) } \right) l_k + 2/\kappa + 1/\kappa \sqrt{ g_1(\kappa) }
\right).
\end{align*}
Since $\left( 2 - \sqrt{ g_1(\kappa) } \right) \leq 0$ (proven in
Lemma~\ref{LemFact3}) and $l_k \leq 1$, we can upper bound $\lambda_k^{-}$
with an expression independent of $l_k$:
\begin{align*}
\lambda_k^{-} \leq - \frac{\kappa}{4} \left( \left( 2 - \sqrt{
  g_1(\kappa) } \right) + 2/\kappa + 1/\kappa \sqrt{ g_1(\kappa) }
\right).
\end{align*}
We state one final auxiliary lemma:
\begin{lemma}
  \label{LemFact3}
  The function $g_1$ given in equation~\eqref{EqnDefnGone} has the following properties:
\begin{subequations} 
  \begin{align}
    \label{eq:fact3_1}
  2 - \sqrt{ g_1(\kappa) } & \leq 0 \qquad \mbox{for all $\kappa \geq
    1$, and} \\
\label{eq:fact3_2}  
- \frac{\kappa}{4} \left( \left( 2 - \sqrt{ g_1(\kappa) } \right) +
2/\kappa + 1/\kappa \sqrt{ g_1(\kappa) } \right) & \leq -1/5.
  \end{align}
\end{subequations}
\end{lemma}
\noindent See Appendix~\ref{appnd:fact3_proof} for the proof of this
claim.  Applying the bound~\eqref{eq:fact3_2} from
Lemma~\ref{LemFact3}, we conclude that $\mathrm{eig}_i(SM_t+M_t^\rT S)
\leq -1/5$ for all $i = 1, \cdots, 3d$, which completes the proof of
Lemma~\ref{lemma:eig}.


\subsubsection{Proof of Lemma~\ref{lemma:eig_S}}

We show that the eigenvalues $\mathrm{eig}_i(S)$ satisfy a certain
third-order equation.  For each \mbox{$i = 1, \ldots, 3d$,} we claim
that the variable $x = \frac{4}{\kappa^5} \cdot \mathrm{eig}_i(S)$,
satisfies the following cubic equation:
\begin{equation}
\label{eq:auxiliary_equation}  
f(x) = x^3 - g_2(\kappa) \cdot x^2 + 1/\kappa^9 g_3(\kappa) \cdot x -
1/\kappa^{15} g_3(\kappa) = 0,
\end{equation}
where the coefficients $g_2(\kappa)$ and $g_3(\kappa)$ are defined in
Appendix~\ref{appnd:g}.  Since $S$ is a symmetric matrix, all the
roots of equation~\eqref{eq:auxiliary_equation} are real.

In order for the eigenvalues $\mathrm{eig}_i(S)$ to lie in the
interval $[\frac{1}{5\kappa}, \kappa^2 + 10]$, it suffices to show
that the roots of the function $f$ from
equation~\eqref{eq:auxiliary_equation} all lie in the interval $[
  \frac{4}{5\kappa^6}, \; \frac{4}{\kappa^3} + \frac{40}{\kappa^5}]$.

\begin{lemma}
  \label{lemma:f_bounds}
  For any $\kappa\geq1$, $f(x)$ defined in
  equation~\eqref{eq:auxiliary_equation} satisfy that $\forall x \leq
  \frac{4}{5\kappa^6}$, $f(x) < 0$, and that $\forall x \geq
  \frac{4}{\kappa^3} + \frac{40}{\kappa^5}$, $f(x) > 0$.
\end{lemma}

\noindent We defer the proof of this lemma to
Appendix~\ref{appnd:lemma_f_bounds}.  Note that
Lemma~\ref{lemma:f_bounds} implies that all real roots of
equation~\eqref{eq:auxiliary_equation} lie in the range of $\left[
  \frac{4}{5\kappa^6}, \frac{4}{\kappa^3} + \frac{40}{\kappa^5}
  \right]$, which completes the proof.


\subsection{Proofs of discrete-time results}
\label{SecDiscreteTime}

We now turn to the proofs of our two main results---namely,
Theorems~\ref{thm:finite-sum-main} and
Theorem~\ref{thm:abstract-highly-smooth}---that establish the behavior of the
discrete-time algorithm.  We begin with a general roadmap for the
proofs, along with a key auxiliary result
(Proposition~\ref{prop:one-step-discr}) common to both arguments.


\subsubsection{Roadmap and a key auxiliary result}
\label{SecRoadmap}

Let $(\discretized{x}_t = (\discretized{\theta}_t,
\discretized{p}_t, \discretized{r}_t) : t \geq 0)$ be the process
defined in Section~\ref{sec:implementation}. For $K \in \mathbb{N}$,
$\discretized{x}_{k \stepsize}$ has the same distribution as $x^{(k)}$
defined by Algorithm~\ref{alg:main}. We construct a coupling between
the process $\discretized{x}$ and the stationary diffusion process
$(x_t : t \geq 0)$ with $x_0 \sim e^{- U (\theta) -
  \frac{L}{2}\vecnorm{p}{2}^2 - \frac{L}{2} \vecnorm{r}{2}^2}$. (It is
obvious that $x_t$ is also following the stationary distribution for
any $t > 0$). Given a $d$-dimensional Brownian motion, $(B_t: t \geq
0)$, we use it to drive both processes. For any $N \in \mathbb{N}$,
we have:
\begin{align*}
  \mathcal{W}_2 (\pi^{(N)}, \pi) \leq \left( \Exs
  \vecnorm{\discretized{x}_{N \stepsize} - x_{N \stepsize}}{2}^2
  \right) \leq \matsnorm{S^{-1}}{op} \left( \Exs
  \vecnorm{\discretized{x}_{N \stepsize} - x_{N \stepsize}}{S}^2
  \right).
\end{align*}
In the last step we transform the $\vecnorm{\cdot}{2}$ norm
into the $\vecnorm{\cdot}{S}$ norm, which is possible because the process $(x_t : t \geq 0)$
is contractive under the $\vecnorm{\cdot}{S}$ norm.

To analyze the one-step discretization error, we have the following key lemma,
which holds in general for any approximator of $\nabla U$.  Note that
Proposition~\ref{prop:one-step-discr} is used in the proof of both
theorems.

\begin{proposition}
  \label{prop:one-step-discr}
Given an $(m,L)$-convex-smooth potential, let the process
$\discretized{x}_t = (\discretized{\theta}_t, \discretized{p}_t
\discretized{r}_t)$ be defined by
equations~\eqref{eq:underdamp-euler}--\eqref{eq:underdamp-correction-general},
and let $x_t = (\theta_t, p_t, r_t)$ be generated from the
continuous-time dynamics equation~\eqref{eq:3order_dyn} initialized
with the stationary distribution. Suppose that we use the same
Brownian motion for both processes, and assume that $\hat{g}_t
(\theta_1, \theta_2)$ belongs to $\mathrm{conv} \left( \left\{ \nabla
U (\lambda \theta_1 + (1 - \lambda) \theta_2): \lambda \in [0, 1]
\right\} \right)$.  Under these conditions, there is a universal
constant $C$ such that
\begin{subequations}
\begin{multline}
  \Exs \vecnorm{\discretized{x}_{(k + 1) \stepsize} - x_{(k + 1)
      \stepsize}}{S}^2 \leq \left( 1 - \frac{ \stepsize }{20\kappa^2 +
    200} \right) \Exs\vecnorm{\discretized{x}_{k \stepsize} - x_{k
      \stepsize}}{S}^2 + C \frac{\kappa^{8} \stepsize^5 d}{L} \\
  + C \frac{\kappa^5 \stepsize \Exs \Delta_k (g)^2}{L^2} + C \kappa^6
  \frac{\stepsize^5}{L^2} \Exs \sup_{k \stepsize \leq s \leq (k + 1)
    \stepsize} \vecnorm{\frac{\rd}{\rd s} \hat{g}_s
    \left(\discretized{\theta}_{k \stepsize}, \hat{\theta}_{(k + 1)
      \stepsize}\right)}{2}^2,
\end{multline}
where
\begin{align}
  \Delta_k (g) \mydefn \sup_{k \stepsize \leq t \leq (k + 1)
    \stepsize} \vecnorm{\hat{g}_{t} \left( \discretized{\theta}_{k
      \stepsize}, \hat{\theta}_{(k + 1) \stepsize} \right) - \nabla U
    \left( \frac{t - k \stepsize}{ \stepsize} \hat{\theta}_{(k + 1)
      \stepsize} + \frac{(k + 1) \stepsize - t}{\stepsize}
    \tilde{\theta}_{k \stepsize} \right) }{2}.
\end{align}
\end{subequations}
\end{proposition}

Note that the process $\discretized{x}_t$ also satisfies a system of
SDEs, with drift terms defined by $\hat{g}, \hat{p}$ and
$\hat{r}$. Those drift terms are dependent on the future moves of
the Brownian motion, so the equation for the interpolation of
discrete-time process is not a Markov diffusion. Nevertheless, we can
still compare it with the process~\eqref{eq:3order_dyn} using a
synchronous coupling, and compute the evolution of
$\vecnorm{\discretized{x}_t - x_t}{S}^2$ along the path:
\begin{align*}
    \vecnorm{\discretized{x}_t - x_t}{S}^2 \leq &
    \vecnorm{\discretized{x}_{k \stepsize} - x_{k \stepsize}}{S}^2 -
    \frac{1}{20 \kappa^2 + 200} \int_{k \stepsize}^t
    \vecnorm{\discretized{x}_s - x_s}{S}^2 ds\\ &+ (20 \kappa^2 + 200)
    \matsnorm{S}{op} \int_{k \stepsize}^t \left( \vecnorm{\hat{g}_s -
      \nabla U (\discretized{\theta}_s)}{2}^2 - \vecnorm{\hat{r}_s -
      \discretized{r}_s}{2}^2 + (1 + \gamma^2) \vecnorm{\hat{p}_s -
      \discretized{p}_s}{2}^2 \right) \rd s.
\end{align*}
Comparing the constructions in
equations~\eqref{eq:underdamp-euler}--\eqref{eq:underdamp-correction-general},
we note that:
\begin{align*}
    \hat{\theta}_t - \discretized{\theta}_t = & \int_{k \stepsize}^t
    \int_{k \stepsize}^s (- \hat{g}_\ell / L - \gamma \hat{r}_\ell)
    \rd \ell \rd s,\\
    \hat{r}_t - \discretized{r}_t = & \gamma \int_{k
      \stepsize}^t \int_{k \stepsize}^s (- \hat{g}_\ell / L - \gamma
    \hat{r}_\ell ) \rd\ell \rd s - \xi \int_{k \stepsize}^t (\hat{r}_s - \discretized{r}_s) \rd s,
\end{align*}

where the $O(\sqrt{d})$ difference, caused by the difference between
drifts taken at discrete-time and continuous-time points, is
integrated twice. The term $\hat{p}_t - \discretized{p}_t$ is
actually the error for approximation of the path using a linear
function. Since $\hat{p}_t$ is an integral by itself, we gain one more
order of smoothness: the process of interest has a third-order time
derivative bounded by $O(\sqrt{d})$, which is then integrated twice in
the final bound. This leads to an $O(\sqrt{\stepsize^3 d})$ bound on
the bias, making it possible for the rate to go below
$O(\sqrt{d})$.


\subsubsection{Proof of Theorem~\ref{thm:finite-sum-main}}

We now turn to the proof of Theorem~\ref{thm:finite-sum-main}. Let
$\hat{g}_t (\theta_1, \theta_2) = \nabla U \left( \frac{(k + 1)
  \stepsize - t}{\stepsize} \theta_1 + \frac{t - k
  \stepsize}{\stepsize} \theta_2 \right)$. Since the function $U$ is
in the form of equation~\eqref{eq:form-generalized-linear}, by
Section~\ref{sec:implementation}, the one-step update can be
explicitly solved in closed form. We thus have $\Delta_k (g) = 0$ for any
$k \in \mathbb{N}_+$.

By Proposition~\ref{prop:one-step-discr}, for the synchronous coupling
between the process $\discretized{x}_t$ defined by the algorithm and
the process~\eqref{eq:3order_dyn}, we have
\begin{align*}
  \Exs \vecnorm{\discretized{x}_{(k + 1) \stepsize} - x_{(k + 1)
      \stepsize}}{S}^2 &\leq \left( 1 - \frac{ \stepsize }{20\kappa^2 +
    200} \right) \Exs\vecnorm{\discretized{x}_{k \stepsize} - x_{k
      \stepsize}}{S}^2 + C  \kappa^{8} \stepsize^5 d / L \\
&\quad \quad  + C \kappa^5 \stepsize \Exs \Delta_k (g)^2 / L^2 + C \kappa^6
  \frac{\stepsize^5}{L^2} \Exs \sup_{k \stepsize \leq s \leq (k + 1)
    \stepsize} \vecnorm{\frac{\rd}{\rd s} \hat{g}_s
    \left(\discretized{\theta}_{k \stepsize}, \hat{\theta}_{(k + 1)
      \stepsize}\right)}{2}^2 \\ 
      &\leq \left( 1 - \frac{ \stepsize
  }{20\kappa^2 + 200} \right) \Exs\vecnorm{\discretized{x}_{k
      \stepsize} - x_{k \stepsize}}{S}^2 +  C  \kappa^{8} \stepsize^5 d / L\\
      &\quad \quad + C
  \kappa^6 \frac{\stepsize^5}{L^2} \Exs \sup_{k \stepsize \leq s \leq
    (k + 1) \stepsize} \vecnorm{\frac{\rd}{\rd s} \nabla U \left(
    \discretized{\theta}_{k \stepsize} + (s - k \stepsize)
    \discretized{p}_{k \stepsize} \right) }{2}^2.
\end{align*}
Note that by the $L$-smoothness condition (Assumption~\ref{AssAll}),
we have
\begin{align*}
  \vecnorm{\frac{\rd}{\rd s} \nabla U \left(
    \discretized{\theta}_{k \stepsize} + (s - k \stepsize)
    \discretized{p}_{k \stepsize} \right) }{2} &= \vecnorm{
    \nabla^2 U \left( \discretized{\theta}_{k \stepsize} + (s -
    k \stepsize) \discretized{p}_{k \stepsize} \right)
    \tilde{p}_{k \stepsize} }{2}\\ &\leq \matsnorm{\nabla^2 U
    \left( \discretized{\theta}_{k \stepsize} + (s - k
    \stepsize) \discretized{p}_{k \stepsize} \right)}{op} \cdot
  \vecnorm{\discretized{p}_{k \stepsize} }{2} \leq L
  \vecnorm{\discretized{p}_{k \stepsize}}{2}.
\end{align*}
The moments can further be controlled as:
\begin{align*}
  \Exs \vecnorm{\discretized{p}_{k \stepsize}}{2}^2 \leq 2 \Exs
  \vecnorm{p_{k \stepsize}}{2}^2 + 2 \Exs \vecnorm{p_{k
      \stepsize} - \discretized{p}_{k \stepsize}}{2}^2 \leq
  \frac{2 d}{L} + 2 \matsnorm{S^{-1}}{op} \Exs \vecnorm{x_{k
      \stepsize} - \discretized{x}_{k \stepsize}}{S}^2.
\end{align*}
Therefore, with $\stepsize < c \kappa^{- 11/ 4}$, we have $2 C
\kappa^6 \stepsize^5 \matsnorm{S^{-1}}{op} \leq \frac{\stepsize}{40
  \kappa^2 + 400}$, and consequently:
\begin{align*}
  \Exs \vecnorm{\discretized{x}_{(k + 1) \stepsize} - x_{(k + 1)
      \stepsize}}{S}^2 &\leq \left( 1 - \frac{ \stepsize }{20\kappa^2 +
    200} \right) \Exs\vecnorm{\discretized{x}_{k \stepsize} - x_{k
      \stepsize}}{S}^2  + C
 \kappa^8 \stepsize^5 d / L. \\
&\quad \quad + C \kappa^6 \stepsize^5 \Exs \vecnorm{\discretized{p}_{k
    \stepsize}}{2}^2 \\
&\leq \left( 1 - \frac{ \stepsize }{40\kappa^2 + 400} \right)
\Exs\vecnorm{\discretized{x}_{k \stepsize} - x_{k \stepsize}}{S}^2 + C
 \kappa^8 \stepsize^5 d / L.
\end{align*}
Solving the recursion, we obtain:
\begin{align}
  \Exs \vecnorm{\discretized{x}_{(k + 1) \stepsize} - x_{(k + 1)
      \stepsize}}{S}^2 \leq \left( 1 - \frac{ \stepsize
  }{20\kappa^2 + 200} \right)^k \Exs \vecnorm{\discretized{x}_0
    - x_0}{S}^2 + C'  \kappa^{10} \stepsize^4 d / L. \label{eq:glm-bound-final}
\end{align}
Starting the algorithm from $\discretized{x}_0 \mydefn
(\theta_0, 0, 0)$, we have $\Exs \vecnorm{x_0 -
  \discretized{x}_0}{2}^2 \leq \frac{8 d}{L}$.

For a given $\varepsilon > 0$, to make the desired bound hold, we
use the synchronous coupling we constructed, and needing $\Exs
\vecnorm{\discretized{x}_{N \stepsize} - x_{N \stepsize}}{2}^2
\leq \varepsilon^2$, we let both terms in
equation~\eqref{eq:glm-bound-final} scale as at most $\frac{1}{2}
\matsnorm{S^{-1}}{op}^{-1} \varepsilon^2$, which leads to:
\begin{align*}
  \begin{cases}
    \stepsize < c \kappa^{- \frac{11}{4}} d^{- \frac{1}{4}} \varepsilon^{\frac{1}{2}},  \\
    N \geq C \cdot \frac{\kappa^2}{\stepsize} \log \frac{3 d}{L
      \varepsilon}.
  \end{cases}
\end{align*}
Choosing the parameters accordingly completes the proof.

\subsubsection{Proof of Theorem~\ref{thm:abstract-highly-smooth}}

Now we describe the proof of Theorem~\ref{thm:abstract-highly-smooth},
a general result for functions with high-order smoothness. The proof
is also based on synchronous coupling.  In addition to exploiting
Proposition~\ref{prop:one-step-discr}, we also use the following
standard result for Chebyshev node
interpolation~\cite{stewart1998afternotes}:
\begin{lemma}
  \label{lemma:gauss-lengendre}
For a curve $(x_t: 0 \leq t \leq \ell)$ in $\real^d$, let $(\Phi(t; x)
\mid 0 \leq t \leq \ell)$ be the $(\alpha - 1)$-order Lagrange
polynomial defined at the $\alpha$-th order Chebyshev nodes.  Then the
interpolation error is bounded as
  \begin{align}
    \sup_{0 \leq t \leq \ell} \vecnorm{x_t - \Phi(t; x) }{2} \leq
    \frac{1}{2^{\alpha - 1} \alpha !}\ell^\alpha \sup_{0 \leq t \leq
      \ell} \vecnorm{\frac{\rd^{\alpha}}{\rd t^\alpha}{x}_t}{2}.
  \end{align}
\end{lemma}

\noindent Now suppose that we use $(\alpha - 1)$-th order Chebyshev
nodes in order to construct a polynomial $\hat{g}_t$ that approximates
$\nabla U (\discretized{\theta}_{k \stepsize} + (t - k \stepsize)
\discretized{p}_{k \stepsize})$. By Lemma~\ref{lemma:gauss-lengendre},
the approximation error can be controlled by:
\begin{align*}
    \Delta_k (g) &= \sup_{k \stepsize \leq t \leq (k + 1)
      \stepsize}\vecnorm{\hat{g}_t - \nabla U (\discretized{\theta}_{k
        \stepsize} + (t - k \stepsize) \discretized{p}_{k
        \stepsize})}{2}\\ &\leq  \stepsize^{\alpha - 1} \sup_{k
      \stepsize \leq t \leq (k + 1) \stepsize}
    \vecnorm{\frac{\rd^{\alpha - 1}}{\rd s^{\alpha - 1}} \nabla U
      (\discretized{\theta}_{k \stepsize} + (s - k \stepsize)
      \discretized{p}_{k \stepsize} }{2} \\ &\leq \stepsize^{\alpha -
      1} \sup_{k \stepsize \leq t \leq (k + 1) \stepsize}
    \matsnorm{\nabla^\alpha U (\discretized{\theta}_{k \stepsize} + (t
      - k \stepsize) \discretized{p}_{k \stepsize}}{\tens}^{(\alpha)}
    \cdot \vecnorm{\discretized{p}_{k \stepsize}}{2}^{\alpha -
      1}\\ &\leq \stepsize^{\alpha - 1} L_\alpha^{\alpha - 1} \cdot
    \vecnorm{\discretized{p}_{k \stepsize}}{2}^{\alpha - 1}.
\end{align*}
So by Lemma~\ref{lemma:high-moments}, we have:
\begin{align*}
    \Exs \Delta_k (g)^2 \leq \left( C \stepsize L_{\alpha} \kappa^3
    \sqrt{\frac{d + 2 \alpha}{L}} \right)^{2(\alpha - 1)}.
\end{align*}
For Lagrange interpolating polynomials, the time derivative is the
finite difference between interpolation points. These differences can
be further bounded by the time derivative of the original process
$\nabla U (\theta + t p)$---viz.
\begin{align*}
 \sup_{0 \leq s \leq \stepsize} \vecnorm{\frac{\rd}{\rd s}\hat{g}_{k
     \stepsize + s}}{2} \leq L \vecnorm{\discretized{p}_{k
     \stepsize}}{2}.
\end{align*}

Similar to the proof of Theorem~\ref{thm:finite-sum-main}, since the
weights in Lagrangian interpolation at Chebyshev nodes are
non-negative, using Proposition~\ref{prop:one-step-discr}, we obtain
the bound:
\begin{align*}
  \vecnorm{\discretized{x}_{(k + 1) \stepsize} - x_{(k + 1)
      \stepsize}}{2}^2 \leq \left( 1 - \frac{ \stepsize }{40\kappa^2 +
    400} \right) \Exs\vecnorm{\discretized{x}_{k \stepsize} - x_{k
      \stepsize}}{S}^2 + C \kappa^{8} \stepsize^5 d / L  + \stepsize \left( C \stepsize L_{\alpha} \kappa^3
  \sqrt{\frac{d + 2 \alpha}{L}} \right)^{2(\alpha - 1)}.
\end{align*}

In order to ensure that $\Exs \vecnorm{\discretized{x}_{N \stepsize} -
  x_{N \stepsize}}{2}^2 \leq \varepsilon^2$, we need each of the
three error terms to be bounded by $\frac{\varepsilon}{3}$.  Solving
the resulting equations leads to
\begin{align*}
  \begin{cases}
    \stepsize < c \cdot \min \left( \kappa^{- \frac{11}{4}}
    d^{ - \frac{1}{4} } L^{- \frac{1}{4}} \varepsilon^{-
      \frac{1}{2}},
    L_{\alpha}^{-1} \kappa^{- 3 - \frac{1}{\alpha - 1}}
            L^{\frac{1}{2}} d^{- \frac{1}{2}} \varepsilon^{-
              \frac{1}{\alpha - 1}} \right) \\ N \geq C \cdot
            \frac{\kappa^2}{\stepsize} \log \frac{3 d}{L \varepsilon},
  \end{cases}
\end{align*}
which completes the proof.



\section{Discussion}
\label{SecDiscussion}

In this paper, we focus on accelerating the convergence of
gradient-based MCMC algorithms in high-dimensional spaces. We break the problem into two parts:
a splitting scheme that reduces the problem of SDE discretization to that of integration along a fixed straight line;
a third-order Langevin dynamics in continuous time which allows this fine-grained discretization analysis while satisfying exponentially fast convergence.


For the second problem, we construct a third-order Langevin dynamics so that the
trajectories are smoother and the integration of $\nabla U$ is separated from the Brownian motion part. We then apply utilize this dynamics to design MCMC algorithms adaptive to underlying structures of the problem. A mixing time of order $\mathcal{O}\left(d^{1/4} / \varepsilon^{1/2}\right)$ is achieved for ridge-separable potentials, which cover a large class of machine learning models. Under black-box oracle model, a rate of order $O(d^{1/4} /
\varepsilon^{1/2} + d^{1/2} / \varepsilon^{1/(\alpha - 1)} )$ is
achieved for $k$-th order smooth objective functions $U$.

An important future direction is to further investigate higher-order splitting scheme with the use of higher-order dyanmics, to further reduce the dimension dependency for the mixing time of MCMC. We conjecture that the exponent on $d$ can be further reduced, with a trade-off between the dependency on the dimension and condition number.

\subsection*{Acknowledgements} This work was partially supported by Office of
Naval Research Grant ONR-N00014-18-1-2640, Army Research Office grant
W911NF-17-1-0304, and National Science Foundation Grants
NSF-IIS-1740855, NSF-IIS-1909365, and NSF-IIS-1619362. We thank Santosh Vempala and Chris Junchi Li for helpful discussions.


\bibliography{CompleteSampling}

\begin{thebibliography}{10}

\bibitem{Splitting}
A.~Abdulle, G.~Vilmart, and K.~C. Zygalakis.
\newblock {Long time accuracy of Lie\mbox{-}Trotter splitting methods for
  Langevin dynamics}.
\newblock {\em SIAM J. Numer. Anal.}, 53(1):1--16, 2015.

\bibitem{Eberle_HMC}
N.~Bou-Rabee, A.~Eberle, and R.~Zimmer.
\newblock {Coupling and convergence for Hamiltonian Monte Carlo}.
\newblock arXiv:1805.00452, 2018.

\bibitem{Opt_Dyn}
A.~A. Brown and M.~C. Bartholomew-Biggs.
\newblock Some effective methods for unconstrained optimization based on the
  solution of systems of ordinary differential equations.
\newblock {\em J. Optimiz. Theory App.}, 62(2):211--224, 1989.

\bibitem{cao2020complexity}
Y.~Cao, J.~Lu, and L.~Wang.
\newblock Complexity of randomized algorithms for underdamped langevin
  dynamics.
\newblock {\em arXiv preprint arXiv:2003.09906}, 2020.

\bibitem{Variance_Reduction_theory}
N.~Chatterji, N.~Flammarion, Y.-A. Ma, P.~Bartlett, and M.~Jordan.
\newblock On the theory of variance reduction for stochastic gradient {M}onte
  {C}arlo.
\newblock In {\em Proceedings of the 35th International Conference on Machine
  Learning (ICML)}, volume~80, pages 764--773, 2018.

\bibitem{Che+19_HMC}
Y.~Chen, R.~Dwivedi, M.~J. Wainwright, and B.~. Yu.
\newblock Fast mixing of {M}etropolized {H}amiltonian {M}onte {C}arlo:
  {B}enefits of multi-step gradients.
\newblock Technical report, UC Berkeley, May 2019.
\newblock arXiv:1905.12247.

\bibitem{Xiang_overdamped}
X.~Cheng and P.~L. Bartlett.
\newblock Convergence of {L}angevin {MCMC} in {KL}-divergence.
\newblock In {\em Proceedings of the 29th International Conference on
  Algorithmic Learning Theory (ALT)}, pages 186--211, 2018.

\bibitem{Xiang_Nonconvex}
X.~Cheng, N.~S. Chatterji, Y.~Abbasi-Yadkori, P.~L. Bartlett, and M.~I. Jordan.
\newblock Sharp convergence rates for {L}angevin dynamics in the nonconvex
  setting.
\newblock arXiv:1805.01648, 2018.

\bibitem{Xiang_underdamped}
X.~Cheng, N.~S. Chatterji, P.~L. Bartlett, and M.~I. Jordan.
\newblock Underdamped {Langevin MCMC}: A non-asymptotic analysis.
\newblock In {\em Proceedings of the 31st Conference on Learning Theory
  (COLT)}, pages 300--323, 2018.

\bibitem{Dalalyan_JRSSB}
A.~S. Dalalyan.
\newblock Theoretical guarantees for approximate sampling from smooth and
  log-concave densities.
\newblock {\em J. Royal Stat. Soc. B}, 79(3):651--676, 2017.

\bibitem{Dalalyan_user_friendly}
A.~S. Dalalyan and A.~G. Karagulyan.
\newblock User-friendly guarantees for the {Langevin Monte Carlo} with
  inaccurate gradient.
\newblock {\em Stochastic Process. Appl.}, 2019.

\bibitem{Dalalyan_underdamped}
A.~S. Dalalyan and L.~Riou-Durand.
\newblock On sampling from a log-concave density using kinetic {L}angevin
  diffusions.
\newblock arXiv:1807.09382, 2018.

\bibitem{Moulines_ULA}
A.~Durmus and E.~Moulines.
\newblock {High-dimensional Bayesian inference via the Unadjusted Langevin
  Algorithm}.
\newblock arXiv:1605.01559v4, 2018.

\bibitem{dwivedi2018log}
R.~Dwivedi, Y.~Chen, M.~J. Wainwright, and B.~Yu.
\newblock Log-concave sampling: {Metropolis-Hastings} algorithms are fast!
\newblock arXiv:1801.02309, 2018.

\bibitem{Ledoux_log_Sobolev_diffusion}
M.~Ledoux.
\newblock The geometry of {M}arkov diffusion generators.
\newblock {\em Ann Fac Sci Toulouse Math}, 9(6):305--366, 2000.

\bibitem{YinTat}
Y.-T. Lee, Z.~Song, and S.~S. Vempala.
\newblock Algorithmic theory of {ODEs} and sampling from well-conditioned
  logconcave densities.
\newblock arXiv:1812.06243, 2018.

\bibitem{Thermostat_Math}
B.~Leimkuhler and X.~Shang.
\newblock Adaptive thermostats for noisy gradient systems.
\newblock {\em SIAM J. Sci. Comput.}, 38(2):A712--A736, 2016.

\bibitem{Yian_underdamped}
Y.-A. Ma, N.~S. Chatterji, X.~Cheng, N.~Flammarion, P.~L. Bartlett, and M.~I.
  Jordan.
\newblock Is there an analog of {Nesterov} acceleration for {MCMC}?
\newblock arXiv:1902.00996, 2019.

\bibitem{completesample}
Y.-A. Ma, T.~Chen, and E.~B. Fox.
\newblock A complete recipe for stochastic gradient {MCMC}.
\newblock In {\em Advances in Neural Information Processing Systems 28 (NIPS)},
  pages 2899--2907. 2015.

\bibitem{MCMC_nonconvex}
Y.-A. Ma, Y.~Chen, C.~Jin, N.~Flammarion, and M.~I. Jordan.
\newblock Sampling can be faster than optimization.
\newblock arXiv:1811.08413, 2018.

\bibitem{completeframework}
Y.-A. Ma, E.~B. Fox, T.~Chen, and L.~Wu.
\newblock Irreversible samplers from jump and continuous {M}arkov processes.
\newblock {\em Stat. Comput.}, pages 1--26, 2018.

\bibitem{Mangoubi1}
O.~Mangoubi and A.~Smith.
\newblock Rapid mixing of {Hamiltonian Monte Carlo} on strongly log-concave
  distributions.
\newblock arXiv:1708.07114, 2017.

\bibitem{Mangoubi2}
O.~Mangoubi and N.~K. Vishnoi.
\newblock {Dimensionally tight running time bounds for second-order Hamiltonian
  Monte Carlo}.
\newblock In {\em Advances in Neural Information Processing Systems (NeurIPS)
  32}, pages 6030--6040. 2018.

\bibitem{NealHMC}
R.~M. Neal.
\newblock {MCMC} using {Hamiltonian} dynamics.
\newblock {\em Handbook of {M}arkov Chain {M}onte {C}arlo}, 54:113--162, 2010.

\bibitem{Nesterov_intro}
Y.~Nesterov.
\newblock {\em Introductory Lectures on Convex Optimization: A Basic Course}.
\newblock Kluwer, Boston, 2004.

\bibitem{Gareth}
G.~O. Roberts and J.~S. Rosenthal.
\newblock Optimal scaling for various {Metropolis-Hastings} algorithms.
\newblock {\em Statist. Sci.}, 16(4):351--367, 2001.

\bibitem{shen2019randomized}
R.~Shen and Y.~T. Lee.
\newblock The randomized midpoint method for log-concave sampling.
\newblock In {\em Advances in Neural Information Processing Systems}, pages
  2098--2109, 2019.

\bibitem{shi2018understanding}
B.~Shi, S.~S. Du, M.~I. Jordan, and W.~J. Su.
\newblock Understanding the acceleration phenomenon via high-resolution
  differential equations.
\newblock {\em arXiv preprint arXiv:1810.08907}, 2018.

\bibitem{stewart1998afternotes}
G.~W. Stewart.
\newblock {\em Afternotes Goes to Graduate School: Lectures on Advanced
  Numerical Analysis}, volume~58.
\newblock Siam, 1998.

\bibitem{Numerical_Analysis}
J.~Stoer and R.~Bulirsch.
\newblock {\em Introduction to Numerical Analysis}.
\newblock Springer, New York, 3rd edition, 2002.

\bibitem{Weijie}
W.~Su, S.~Boyd, and E.~Candes.
\newblock A differential equation for modeling {N}esterov's accelerated
  gradient method: Theory and insights.
\newblock In Z.~Ghahramani, M.~Welling, C.~Cortes, N.~D. Lawrence, and K.~Q.
  Weinberger, editors, {\em Advances in Neural Information Processing Systems
  27}, pages 2510--2518. 2014.

\bibitem{Villani_optimal_transport}
C.~Villani.
\newblock {\em Optimal Transport: Old and New}.
\newblock Wissenschaften. Springer, Berlin, 2009.

\bibitem{Ashia}
A.~Wilson, B.~Recht, and M.~I. Jordan.
\newblock A {L}yapunov analysis of momentum methods in optimization.
\newblock arXiv:1611.02635, 2016.

\end{thebibliography}

\appendix

\section{Proof of Proposition~\ref{prop:one-step-discr}}
\label{AppProofPropOneStep}

In this appendix, we prove Proposition~\ref{prop:one-step-discr}, as
previously stated in Section~\ref{SecRoadmap}.  Recall that this
result provides a bound on the discretization error with certain
choice of $\hat{g}_t$ used in equation~\eqref{eq:integration-p}. Our
proof of this bound is based on direct coupling estimates.


\subsection{A decomposition into three terms}

Comparing the two processes along the path, we obtain:
\begin{align*}
  \rd \left[\begin{matrix} \theta_s - \discretized{\theta}_s\\ p_s
      - \discretized{p}_s\\ r_s - \discretized{r}_s
    \end{matrix}
    \right] &= (D + Q) \left[ \begin{matrix}
      \nabla U (\theta_s) - \nabla U(\discretized{\theta}_s) \\
      L (p_s - \discretized{p}_s)\\
      L (r_s - \discretized{r}_s)
    \end{matrix}\right] \rd s + 
  \frac{1}{L} \left[
    \begin{matrix}
      0\\ \nabla U(\discretized{\theta}_s) - \hat{g}_{s} \left(
      \discretized{\theta}_{k \stepsize}, \hat{\theta}_{(k + 1)
        \stepsize} \right) \\ 0
    \end{matrix}
    \right] \rd s\\ &+ \left[
    \begin{matrix}
      0 \\
      \hat{r}_s - \discretized{r}_s\\
      0
    \end{matrix}
    \right] \rd s
  + \left[
    \begin{matrix}
      \discretized{p}_{s} - \hat{p}_s\\ 0\\ \gamma (\hat{p}_s -
      \discretized{p}_s)
    \end{matrix}
    \right] \rd s
\end{align*}
Introducing the function $J(u) = u^T S u$, for the one-step analysis
with $t \in [k \stepsize, (k + 1) \stepsize]$, we have:
\begin{multline*}
J(\discretized{x}_t - x_t) =  J(\discretized{x}_{k \stepsize} - x_{k
  \stepsize}) + \int_{k \stepsize}^t \left[\begin{matrix} \theta_s -
    \discretized{\theta}_s \\ p_s - \discretized{p}_s\\ r_s -
    \discretized{r}_s
    \end{matrix}
    \right]^\rT S(D + Q) \left[ \begin{matrix} \nabla U (\theta_s) -
    \nabla U(\discretized{\theta}_s)\\ L (p_s - \discretized{p}_s)\\ L
    (r_s - \discretized{r}_s)
    \end{matrix}\right] \rd s \notag \\
   + \frac{1}{L} \int_{k \stepsize}^t  (\discretized{x}_s - x_s)^\rT S \left[
    \begin{matrix}
      0\\ \nabla U(\discretized{\theta}_s) - \hat{g}_{s}  \\ 0
    \end{matrix}
    \right] \rd s
  +  \int_{k \stepsize}^t (\discretized{x}_s - x_s)^\rT S  \left[
    \begin{matrix}
      0 \\
      \hat{r}_t - \discretized{r}_t\\
      0
    \end{matrix}
    \right] \rd s.
\end{multline*}
Consequently, we have
\begin{multline*}
  J(\discretized{x}_t - x_t)  \leq \vecnorm{\discretized{x}_{k
      \stepsize} - x_{k \stepsize}}{S}^2 - \frac{1}{5\kappa^2 + 50}
  \int_{k \stepsize}^t \vecnorm{\discretized{x}_s - x_s}{S}^2 \rd s +
  \frac{3}{20 \kappa^2 + 200} \int_{k \stepsize}^t
  \vecnorm{\discretized{x}_s - x_s}{S}^2 \rd s \notag \\ + (20
  \kappa^2 + 200) \matsnorm{S}{op} \int_{k \stepsize}^t \left(
  \frac{1}{L^2} \underbrace{\vecnorm{ \nabla U(\discretized{\theta}_s)
      - \hat{g}_{s}}{2}^2}_{I_1 (s)} +
  \underbrace{\vecnorm{ \hat{r}_t - \discretized{r}_t}{2}^2}_{I_2 (s)}
  + (1 + \gamma^2) \underbrace{\vecnorm{\hat{p}_s -
      \discretized{p}_s}{2}^2}_{I_3 (s)} \right) \rd s.
\end{multline*}
Simplifying further, we have shown that
\begin{multline}
\label{eq:S-norm-decomposition}  
    J(\discretized{x}_t - x_t) \leq \vecnorm{\discretized{x}_{k
        \stepsize} - x_{k \stepsize}}{S}^2 - \frac{1}{20\kappa^2 +
      200} \int_{k \stepsize}^t \vecnorm{\discretized{x}_s - x_s}{S}^2
    \rd s \\
    + (20 \kappa^2 + 200) \matsnorm{S}{op} \int_{k \stepsize}^t (I_1
    (s) / L^2 + I_2 (s) + (1 + \gamma^2) I_3 (s) ) ds.
\end{multline}
The remainder of the proof is devoted to bounding the terms
$\{I_j\}_{j=1}^3$.


\subsection{Some auxiliary results}

In order to bound the terms $\{I_j\}_{j=1}^3$, we require a number of
auxiliary results, stated here.  Our first result bounds the error in
using the line $\left( \left(1 - \frac{t}{\ell} \right) x_0 +
\frac{t}{\ell} x_\ell \; : \; t \in [0, \ell] \right)$ to approximate
a curve $(x_t : t \in [0, \ell])$ in $\real^d$.
\begin{lemma}
\label{lemma:trapezoidal-path-control}
The straight-line approximation error of the curve is uniformly
bounded as
\begin{align*}
  \sup_{0 \leq t \leq \ell} \vecnorm{\left(1- \frac{t}{\ell} \right) x_0
    + \frac{t}{\ell} x_\ell - x_t}{2} & \leq \ell^2
  \vecnorm{\ddot{x}_t}{2}.
\end{align*}
\end{lemma}


\noindent Our second auxiliary lemma relates the squared Euclidean
norm of the interpolation process
(cf. equations~\eqref{eq:underdamp-euler}--\eqref{eq:underdamp-correction-general})
with the squared norm at discrete time steps.
\begin{lemma}
\label{lemma-2nd-moment}
For the process $(\discretized{p}_t, \discretized{r}_t,
\discretized{\theta}_t)$ defined by
equations~\eqref{eq:underdamp-euler}--\eqref{eq:underdamp-correction-general},
we have
  \begin{align*}
    \sup_{k \stepsize \leq s \leq (k + 1) \stepsize} \Exs
    \left(\vecnorm{\discretized{r}_s}{2}^2 +
    \vecnorm{\discretized{p}_s}{2}^2 \right) \leq C \left(\Exs
    \vecnorm{\discretized{\theta}_{k \stepsize} - \thetastar}{2}^2 +
    \Exs \vecnorm{\discretized{p}_{k \stepsize} }{2}^2 + \Exs
    \vecnorm{\discretized{r}_{k \stepsize}}{2}^2 + d / L \right),
  \end{align*}
  for some universal constant $C > 0$.
    \end{lemma}

\noindent Our third auxiliary lemma upper bounds the higher order
moments of the stochastic process generated by Algorithm~\ref{alg:main}.
These bounds are useful for controlling certain higher order derivatives
along the path.
\begin{lemma}
  \label{lemma:high-moments}
Assuming that $\hat{g}$ satisfies $\hat{g}_t (\theta_1, \theta_2) \in
\mathrm{conv} \left( \left\{ \nabla U (\lambda \theta_1 + (1 -
\lambda) \theta_2): \lambda \in [0, 1] \right\} \right)$ for any
$\theta_1, \theta_2$, consider the process $x^{(k)} = (\theta^{(k)},
p^{(k)}, r^{(k)})$ defined by Algorithm~\ref{alg:main}, for any
$\alpha \in \mathbb{N}_+$, we have
\begin{align*}
  \left( \Exs \vecnorm{x^{(k)} - (\thetastar, 0, 0)}{2}^{2 \alpha}
  \right)^{\frac{1}{2 \alpha}} \leq C \kappa^3 \sqrt{\frac{d + 2 \alpha}{L}},
\end{align*}
for some universal constant $C > 0$.
\end{lemma}
\noindent We return to prove all of these claims in
Section~\ref{SecThreeAux}.


\subsection{Bounding the three terms}

Taking the auxiliary lemmas as given for now, we now bound each of the
terms $\{I_j\}_{j=1}^3$ in succession.

\subsubsection{Upper bound for $I_1$}\label{subsubsec:I1}

Note that:
\begin{align*}
  \Exs I_1 (s) & \leq 2 \Exs \vecnorm{\nabla U(\discretized{\theta}_t)
    - \nabla U \left( \frac{t - k \stepsize}{ \stepsize}
    \hat{\theta}_{(k + 1) \stepsize} + \frac{(k + 1) \stepsize -
      t}{\stepsize} \tilde{\theta}_{k \stepsize} \right)}{2}^2 + 2
  \Exs \Delta_k (g)^2\\ &\leq 2 L^2 \Exs
  \vecnorm{\discretized{\theta}_t - \frac{t - k \stepsize}{ \stepsize}
    \hat{\theta}_{(k + 1) \stepsize} - \frac{(k + 1) \stepsize -
      t}{\stepsize} \tilde{\theta}_{k \stepsize}}{2}^2 + 2 \Exs
  \Delta_k(g)^2.
  \end{align*}
For the first term, note that:
\begin{align*}
\vecnorm{\discretized{\theta}_t - \left( \frac{t - k \stepsize}{
    \stepsize} \hat{\theta}_{\stepsize} + \frac{ (k + 1) \stepsize -
    t}{\stepsize} \tilde{\theta}_{k \stepsize} \right)} {2} & =
\vecnorm{\int_{k \stepsize}^t \discretized{p}_s \rd s - \int_{k
    \stepsize}^t \frac{\hat{\theta}_{(k + 1)\stepsize} -
    \discretized{\theta}_{k \stepsize}}{\stepsize} \rd t}{2} \\ & =
\vecnorm{\int_{k \stepsize}^t \discretized{p}_s \rd s - \int_{k
    \stepsize}^t \frac{1}{\stepsize} \int_{k \stepsize}^{(k +
    1)\stepsize} \discretized{p}_{k \stepsize} \rd\ell \rd s}{2} \\
& \leq \int_{k \stepsize}^t \vecnorm{\discretized{p}_s -
  \discretized{p}_{k \stepsize}}{2} \rd s.
  \end{align*}
Moreover, by definition, we have $\discretized{p}_s -
\discretized{p}_{k \stepsize} = \int_{k \stepsize}^s (- \hat{g}_\ell /
L - \gamma \hat{r}_\ell ) \rd\ell$.  Applying the Cauchy-Schwartz
inequality, we obtain:
  \begin{align*}
      \Exs \vecnorm{\discretized{p}_s - \discretized{p}_{k
          \stepsize}}{2}^2 &\leq 2 \stepsize \int_{k \stepsize}^s
      \left( \Exs \vecnorm{ \hat{g}_\ell }{2}^2 / L^2 + \gamma^2 \Exs
      \vecnorm{\hat{r}_\ell }{2}^2 \right) \rd\ell \\
      & \leq \frac{2 \stepsize^2}{L^2} \Exs \sup_{k \stepsize \leq s
        \leq (k + 1) \stepsize} \vecnorm{\nabla U
        (\hat{\theta}_s)}{2}^2 + 2 \stepsize^2 \gamma^2 \sup_{k
        \stepsize \leq \ell \leq (k + 1) \stepsize} \Exs
      \vecnorm{\hat{r}_\ell}{2}^2.
  \end{align*}
  Therefore, applying Cauchy-Schwartz to the integral, we arrive at:
  \begin{align*}
      \Exs I_1 (t) &\leq 2 L^2 ( t - k\stepsize) \int_{k \stepsize}^t
      \Exs \vecnorm{\discretized{p}_s - \discretized{p}_{k
          \stepsize}}{2}^2 \rd s + 2 \Exs \Delta_k(g)^2 \\ &\leq 2
      \stepsize^4 \Exs \sup_{k \stepsize \leq s \leq (k + 1)
        \stepsize} \vecnorm{\nabla U (\hat{\theta}_s)}{2}^2 + 2
      \stepsize^4 \gamma^2 L^2 \sup_{k \stepsize \leq \ell \leq (k +
        1) \stepsize} \Exs \vecnorm{\hat{r}_\ell}{2}^2 + 2 \Exs
      \Delta_k (g)^2.
  \end{align*}
  
The second term is easy to control by the properties of OU processes:
\begin{align*}
  \Exs \vecnorm{\hat{r}_t}{2}^2 \leq 3\Exs
  \vecnorm{\discretized{r}_{k \stepsize}}{2}^2 + 3 \stepsize^2 \Exs
  \vecnorm{\discretized{p}_{k \stepsize}}{2}^2 + \frac{6 \xi \stepsize
    d}{L}.
\end{align*}
For the gradient norm term, by Assumption~\ref{AssAll}, we can relate them to
moments of $\discretized{\theta}$ and $\discretized{p}$:
\begin{align*}
  \Exs \sup_{k \stepsize \leq s \leq (k + 1) \stepsize}
  \vecnorm{\nabla U (\hat{\theta}_s)}{2}^2 &\leq L^2 \Exs \sup_{k
    \stepsize \leq s \leq (k + 1) \stepsize} \vecnorm{\hat{\theta}_s -
    \thetastar}{2}^2 \\ &\leq 2 L^2 \left( \Exs \vecnorm{
    \discretized{\theta}_{k \stepsize} - \thetastar}{2}^2 +
  \stepsize^2 \Exs \vecnorm{\discretized{p}_{k \stepsize}}{2}^2
  \right).
\end{align*}

The bound on $\Exs \Delta_k (g)^2$ depends on specific choice of the gradient approximator $g$, which will be discussed in the main proof of Theorem~\ref{thm:finite-sum-main} and Theorem~\ref{thm:abstract-highly-smooth}.


\subsubsection{Upper bound for $I_2$}

Recalling the dynamics of $\discretized{r}$ and $\hat{r}$, we have
\begin{align*}
      \hat{r}_t - \discretized{r}_t &= - \gamma \int_{k \stepsize}^t
      (\discretized{p}_{k \stepsize} - \discretized{p}_{s}) \rd s -
      \xi \int_{k \stepsize}^t (\hat{r}_s -
      \discretized{r}_s) \rd s.
  \end{align*}
  Solving it as an ODE for $\hat{r}_t - \discretized{r}_t$ within time interval $[k \stepsize, (k + 1) \stepsize]$, we obtain:
  \begin{align*}
      \hat{r}_t - \discretized{r}_t = - \gamma \int_{k \stepsize}^t       (\discretized{p}_{k \stepsize} - \discretized{p}_{s}) e^{- \xi (t - s)} \rd s.
  \end{align*}
  By Eq~\eqref{eq:integration-p}, we have:
  \begin{align*}
      \discretized{p}_s - \discretized{p}_{k \stepsize} = - \int_{k \stepsize}^s \hat{g}_\ell / \smooth d \ell + \gamma \int_{k \stepsize}^s \hat{r}_\ell d \ell.
  \end{align*}
  
  Note that we have:
  \begin{align*}
      \Exs \vecnorm{- \hat{g}_\ell / L - \gamma \hat{r}_\ell }{2}^2 \leq 2\Exs \sup_{k \stepsize \leq s \leq (k + 1) \stepsize}\vecnorm{\nabla U (\hat{\theta}_s)}{2}^2 / \smooth^2 + 2 \gamma^2 \sup_{k \stepsize \leq \ell \leq (k + 1 \stepsize)} \Exs \vecnorm{\hat{r}_\ell}{2}^2.
  \end{align*}
  Using Cauchy-Schwartz twice, we obtain:
  \begin{multline*}
      \Exs I_2 (t) = \Exs \vecnorm{\hat{r}_t -
        \discretized{r}_t}{2}^2 = \gamma^2 \Exs \vecnorm{ \int_{k \stepsize}^t       (\discretized{p}_{k \stepsize} - \discretized{p}_{s}) e^{- \xi (t - s)} \rd s}{2}^2
        \leq \gamma^2 \stepsize \int_{k \stepsize}^t \Exs \vecnorm{\discretized{p}_{k \stepsize} - \discretized{p}_{s}}{2}^2 \rd s\\
        \leq \gamma^2 \stepsize \int_{k \stepsize}^t \vecnorm{- \int_{k \stepsize}^s \hat{g}_\ell / \smooth d \ell + \gamma \int_{k \stepsize}^s \hat{r}_\ell d \ell}{2}^2 d s \leq \gamma^2 \stepsize^2 \int_{k \stepsize}^t \int_{k \stepsize}^s \Exs \vecnorm{- \hat{g}_\ell / L - \gamma \hat{r}_\ell }{2}^2 d \ell ds\\
        \leq 2 \gamma^2 \stepsize^4 \left(\Exs \sup_{k \stepsize \leq s \leq (k + 1) \stepsize}\vecnorm{\nabla U (\hat{\theta}_s)}{2}^2 / \smooth^2 +  \gamma^2 \sup_{k \stepsize \leq \ell \leq (k + 1) \stepsize} \Exs \vecnorm{\hat{r}_\ell}{2}^2 \right).
  \end{multline*}
The upper bound involve the expected supremum of squared gradient norm along the path of $\hat{\theta}_s$, which is already obtained in Section~\ref{subsubsec:I1}:
\begin{align*}
  \Exs \sup_{k \stepsize \leq s \leq (k + 1) \stepsize}
  \vecnorm{\nabla U (\hat{\theta}_s)}{2}^2 \leq 2 L^2 \left( \Exs \vecnorm{
    \discretized{\theta}_{k \stepsize} - \thetastar}{2}^2 +
  \stepsize^2 \Exs \vecnorm{\discretized{p}_{k \stepsize}}{2}^2
  \right).
\end{align*}
Putting them together, for $t \in [k \stepsize, (k + 1) \stepsize]$ we have:
\begin{align*}
    \Exs I_2 (t) \leq 4 \gamma^2 \stepsize^4 \left( \Exs \vecnorm{
    \discretized{\theta}_{k \stepsize} - \thetastar}{2}^2 +
  \stepsize^2 \Exs \vecnorm{\discretized{p}_{k \stepsize}}{2}^2 + \gamma^2  \sup_{k \stepsize \leq \ell \leq (k + 1 ) \stepsize} \Exs \vecnorm{\hat{r}_\ell}{2}^2
  \right).
\end{align*}
  
\subsubsection{Upper Bound for $I_3$}

Note that both the process $\discretized{p}_s$ and the process
$\hat{p}_s$ can be directly written as integration:
\begin{align*}
  \discretized{p}_t &= \discretized{p}_{k \stepsize} - \int_{k
    \stepsize}^t \frac{1}{L}\hat{g}_s \rd s + \int_{k \stepsize}^t
  \gamma \hat{r}_s \rd s,\\ \hat{p}_t &= \discretized{p}_{k
    \stepsize} - \frac{t - k \stepsize}{L\stepsize}
  \left(\displaystyle\int_{k \stepsize}^{(k + 1) \stepsize} \hat{g}_s
  \rd s \right)\rd s + \int_{k \stepsize}^t \gamma \hat{r}_s \rd s.
  \end{align*}
By Lemma~\ref{lemma:trapezoidal-path-control}, for the process
$\iota(s) \mydefn \int_{k \stepsize}^{k \stepsize + s} \hat{g}_\ell d
\ell$, there is:
  \begin{align*}
      \lefteqn{\sup_{0 \leq s \leq \stepsize} \vecnorm{\left(1 -
          \frac{s}{\stepsize} \right) \iota(0) + \frac{s}{\stepsize}
          \iota(\stepsize) - \iota (s) }{2}} \\ &\leq \stepsize^2
      \sup_{0 \leq s \leq \stepsize} \vecnorm{\ddot{\iota}_s}{2} =
      \stepsize^2 \sup_{k \stepsize \leq s \leq (k + 1) \stepsize}
      \vecnorm{\frac{\rd}{\rd s} \hat{g}_{k \stepsize + s}
        \left(\discretized{\theta}_{k \stepsize}, \hat{\theta}_{(k +
          1) \stepsize}\right)}{2}.
  \end{align*}
So we have:
\begin{align*}
  \vecnorm{\discretized{p}_t - \hat{p}_t}{2} = \frac{1}{L} \vecnorm{
    \left(1 - \frac{s}{\stepsize} \right) \iota(0) +
    \frac{s}{\stepsize} \iota(\stepsize) - \iota (s) }{2} \leq
  \frac{\stepsize^2}{L} \sup_{0 \leq s \leq \stepsize}
  \vecnorm{\frac{\rd}{\rd s} \hat{g}_s \left(\discretized{\theta}_{k
      \stepsize}, \hat{\theta}_{(k + 1) \stepsize}\right)}{2},
\end{align*}
which is the upper bound for $I_3$.


\subsection{Obtaining the final bound}

If we have $\vecnorm{\discretized{\theta}_t - \thetastar}{2},
\vecnorm{\discretized{p}_t}{2}, \vecnorm{\discretized{r}_t}{2} = O
(\sqrt{d})$, the above upper bound for $I_1(t)$ is of order $O(
\stepsize^4 d)$ and the upper bound for $I_2 (t)$ is of order $O
(\stepsize^4 d)$, making it possible to achieve a final discretization
error of order $O (\stepsize^{2} d^{1/2})$.

To make this intuition precise, we use
Lemma~\ref{lemma-2nd-moment}, which shows that the supremum of second moments of
$\discretized{r}$ and $\discretized{p}$ along the path can be related
to the second moment at time $k \stepsize$:
\begin{align}
  \sup_{k \stepsize \leq s \leq (k + 1) \stepsize} \Exs
  \left(\vecnorm{\discretized{r}_s}{2}^2 +
  \vecnorm{\discretized{p}_s}{2}^2 \right) \leq C \left(\Exs
  \vecnorm{\discretized{\theta}_{k \stepsize} - \thetastar}{2}^2 +
  \Exs \vecnorm{\discretized{p}_{k \stepsize} }{2}^2 + \Exs
  \vecnorm{\discretized{r}_{k \stepsize}}{2}^2 + d / L
  \right). \label{eq:moment-lemma}
\end{align}


Plugging the moment upper bounds in equation~\eqref{eq:moment-lemma}
to the estimates for $I_1$, $I_2$ and $I_3$, for $\stepsize < \min
\left( 1/\gamma, 1/\xi \right)$, we obtain:
\begin{align*} 
  \Exs I_1 (s) &\leq C \kappa^2 \stepsize^4 L^2 \left(\Exs
  \vecnorm{\discretized{\theta}_{k \stepsize} - \thetastar}{2}^2 +
  \Exs \vecnorm{\discretized{p}_{k \stepsize} }{2}^2 + \Exs
  \vecnorm{\discretized{r}_{k \stepsize}}{2}^2 + d / L\right) +
  2 \Exs \Delta_k (g)^2, \\
  \Exs I_2 (s) &\leq C\kappa^4
  \stepsize^4 \left(\Exs \vecnorm{\discretized{\theta}_{k
      \stepsize} - \thetastar}{2}^2 + \Exs
  \vecnorm{\discretized{p}_{k \stepsize} }{2}^2 + \Exs
  \vecnorm{\discretized{r}_{k \stepsize}}{2}^2 + d / L \right),\\ 
  \Exs I_3 (s) &\leq
  \frac{\stepsize^4}{L^2} \Exs \sup_{k \stepsize \leq s \leq (k +
    1) \stepsize}\vecnorm{\frac{\rd}{\rd s} \hat{g}_s
    \left(\discretized{\theta}_{k \stepsize}, \hat{\theta}_{(k +
      1) \stepsize}\right)}{2}^2.
\end{align*}
So we have:
\begin{align*}
  \lefteqn{\Exs (I_1 (s)/ L^2 + I_2 (s))} \\ &\leq C\kappa^4
  \stepsize^4 \left(\Exs \vecnorm{\discretized{\theta}_{k \stepsize} -
    \thetastar}{2}^2 + \Exs \vecnorm{\discretized{p}_{k \stepsize}
  }{2}^2 + \Exs \vecnorm{\discretized{r}_{k \stepsize}}{2}^2 + d / L^2
  \right) + 2 \Exs \Delta_k (g)^2 / L^2 
  \\
  & \leq 2 C\kappa^4 \stepsize^4 \left(\Exs \vecnorm{\theta_{k
      \stepsize} - \thetastar}{2}^2 + \Exs \vecnorm{p_{k \stepsize}
  }{2}^2 + \Exs \vecnorm{r_{k \stepsize}}{2}^2 + \Exs \vecnorm{x_{k
      \stepsize} - \discretized{x}_{k \stepsize}}{2}^2 + d / L
  \right) + \frac{2 \Exs \Delta_k (g)^2}{L^2}\\ &\leq 2 C \kappa^4 \stepsize^4 \Exs
  \vecnorm{x_{k \stepsize} - \discretized{x}_{k \stepsize}}{2}^2 + 2 C
  \kappa^4 \stepsize^4 d / L + 2 \Exs
  \Delta_k (g)^2 / L^2.
  \end{align*}
For $\stepsize < c \kappa^{- \frac{11}{4} }$ with sufficiently small
$c > 0$, there is $2 (15 \kappa^2 + 150) C \matsnorm{S}{op}
\matsnorm{S^{-1}}{op} \kappa^4 \stepsize^4 < \frac{1}{10 \kappa^2 +
  100}$. Plugging back to the upper bound for
$\vecnorm{\discretized{x}_t - x_t}{S}^2$ along the path, for $t \in [k
  \stepsize, (k + 1) \stepsize]$, we obtain:
\begin{align*}
  \Exs \vecnorm{\discretized{x}_t - x_t}{S}^2 &\leq \Exs
  \vecnorm{\discretized{x}_{k \stepsize} - x_{k \stepsize}}{S}^2 -
  \frac{1}{20\kappa^2 + 200} \int_{k \stepsize}^t
  \vecnorm{\discretized{x}_s - x_s}{S}^2 \rd s \\ &+ C \kappa^4 (t
  - k \stepsize) \left( \kappa^4 \stepsize^4 d / L + \Exs \Delta_k (g)^2 / L^2 \right)\\ &+
  C \frac{\stepsize^5}{L^2} \kappa^6 \Exs \sup_{k \stepsize \leq s \leq (k +
    1) \stepsize}\vecnorm{\frac{\rd}{\rd s} \hat{g}_s
    \left(\discretized{\theta}_{k \stepsize}, \hat{\theta}_{(k +
      1) \stepsize}\right)}{2}^2.
\end{align*}
Applying Gr\"{o}nwall's inequality yields
\begin{align*}
  \Exs \vecnorm{ \discretized{x}_{(k + 1) \stepsize}  - x_{(k + 1) \stepsize}}{S}^2 & \leq \left( 1 - \frac{ \stepsize }{10\kappa^2
    + 100} \right) \Exs\vecnorm{\discretized{x}_{k \stepsize} - x_{k
      \stepsize}}{S}^2 \\ & + C \left( \frac{\kappa^{8} \stepsize^5 d }{L} + \frac{\kappa^4 \stepsize}{L^2} \Exs \Delta_k (g)^2 +
  \frac{\kappa^6\stepsize^5}{L^2} \Exs \sup_{k \stepsize \leq s \leq (k + 1)
    \stepsize} \vecnorm{\frac{\rd}{\rd s} \hat{g}_s
    \left(\discretized{\theta}_{k \stepsize}, \hat{\theta}_{(k + 1)
      \stepsize}\right)}{2}^2 \right).
\end{align*}

\subsection{Proof of auxiliary lemmas}
\label{SecThreeAux}

We now return to prove the auxiliary results that were stated and used
in the previous sections---namely,
Lemmas~\ref{lemma:trapezoidal-path-control}, ~\ref{lemma-2nd-moment}
and~\ref{lemma:high-moments}.


\subsubsection{Proof of Lemma~\ref{lemma:trapezoidal-path-control}}

Introducing the shorthand $\lambda \defn t / \ell$, a direct
calculation with the mean value theorem yields:
  \begin{align*} 
(1 - \lambda) x_0 + \lambda x_\ell- x_{\lambda \ell} & = (1 - \lambda)
    \ell \int_0^{\lambda} \dot{x} (\tau \ell) \rd\tau - \lambda \ell
    \int_0^{1 - \lambda} \dot{x} ((1 - \tau) \ell) \rd\tau \\
& = (1 - \lambda) \ell \lambda \left(\frac{1}{\lambda}
    \int_0^{\lambda} \dot{x} (\tau \ell) \rd\tau - \frac{1}{1 -
      \lambda} \int_0^{1 - \lambda} \dot{x} ((1 - \tau) \ell) \rd\tau
    \right) \\
    & = \lambda (1 - \lambda) \ell \int_0^1 \left( \dot{x} (\tau \ell
    \lambda) - \dot{x} ((1 - (1 - \lambda) \tau) \ell) \right) \rd
    \tau.
  \end{align*}
Taking the Euclidean norm yields
\begin{align*}
  \vecnorm{(1 - \lambda) x_0 + \lambda x_\ell- x_{\lambda \ell}}{2}
  \leq \ell^2 \lambda (1 - \lambda) \sup_{0 \leq t \leq \ell}
  \vecnorm{\ddot{x}_t}{2}.
\end{align*}
    

\subsubsection{Proof of Lemma~\ref{lemma-2nd-moment}}

  For $\discretized{p}_t$, note that:
  \begin{align*}
      \Exs \vecnorm{\discretized{p}_t}{2}^2 \leq 3 \Exs
      \vecnorm{\discretized{p}_{k \stepsize}}{2}^2 + 3 \stepsize
      \int_{k \stepsize}^t \Exs \vecnorm{\hat{g}_s}{2}^2 / L^2 \rd s +
      3 \gamma^2 \stepsize \int_{k \stepsize}^t \Exs
      \vecnorm{\hat{r}_s}{2}^2 \rd s.
  \end{align*}
  The two terms appearing in the above upper bound are both easy to control:
  \begin{align*}
      \Exs \vecnorm{\hat{g}_s}{2}^2 &\leq \Exs \sup_{k \stepsize \leq
        s \leq (k + 1) \stepsize} \vecnorm{\nabla U
        (\hat{\theta}_s)}{2}^2 \leq 2 L^2 \left( \Exs \vecnorm{
        \discretized{\theta}_{k \stepsize} - \thetastar}{2}^2 +
      \stepsize^2 \Exs \vecnorm{\discretized{p}_{k \stepsize}}{2}^2
      \right).\\ 
      \Exs \vecnorm{\hat{r}_s}{2}^2 &\leq 3 \Exs
      \vecnorm{\hat{r}_{k \stepsize}}{2}^2 + 3 \gamma^2 \stepsize
      \int_{k \stepsize}^t \Exs \vecnorm{\discretized{p}_{k
          \stepsize}}{2}^2 dt + 3 \Exs
      \vecnorm{\int_{k \stepsize}^t \sqrt{2 \xi/ L} \rd
        B_t^r}{2}^2\\ &\leq 3 \Exs \vecnorm{\hat{r}_{k
          \stepsize}}{2}^2 + 3 \gamma^2 \stepsize^2 \Exs
      \vecnorm{\discretized{p}_{k \stepsize}}{2}^2 +
      3 \stepsize \xi d / L.
  \end{align*}
  Putting them together, with $\stepsize < \min \left( 1/\gamma, 1/\xi
  \right)$, we have:
  \begin{align*}
      \sup_{k \stepsize \leq t \leq (k + 1) \stepsize} \Exs
      \vecnorm{\discretized{p}_t}{2}^2 &\leq 3 \Exs
      \vecnorm{\discretized{p}_{k \stepsize}}{2}^2 + 6 \stepsize^2
      \left( \Exs \vecnorm{ \discretized{\theta}_{k \stepsize} -
        \thetastar}{2}^2 + \stepsize^2 \Exs
      \vecnorm{\discretized{p}_{k \stepsize}}{2}^2 \right) \\ 
      &\quad \quad + 3
      \gamma^2 \stepsize^2 \left(\stepsize^2 \Exs
      \vecnorm{\discretized{p}_{k \stepsize}}{2}^2 +
      4 \stepsize \xi d / L \right) \\ &\leq 12 \left(\Exs
      \vecnorm{\discretized{\theta}_{k \stepsize} - \thetastar}{2}^2 +
      \Exs \vecnorm{\discretized{p}_{k \stepsize} }{2}^2 + \Exs
      \vecnorm{\discretized{r}_{k \stepsize}}{2}^2 + d / L \right).
  \end{align*}
  
  For $\discretized{r}_t$, the argument is a bit more involved, since
  we need to relate back to the integral of its own moments along the
  path. But since the time interval is short, we can control it
  easily.
  \begin{align*}
    \Exs \vecnorm{\tilde{r}_s - \tilde{r}_{k \stepsize}}{2}^2 &=
    \int_{k \stepsize}^s \left( - \Exs \inprod{\discretized{r}_\ell -
      \discretized{r}_{k \stepsize}}{\discretized{p}_\ell } - \Exs
    \inprod{\discretized{r}_\ell - \discretized{r}_{k
        \stepsize}}{\tilde{r}_\ell} + 2 \xi d / L \right) \rd \ell
    \\ &\leq \frac{1}{2} \int_{k \stepsize}^s \left( \Exs
    \vecnorm{\discretized{r}_\ell - \discretized{r}_{k
        \stepsize}}{2}^2 + \Exs \vecnorm{\discretized{p}_\ell}{2}^2
    \right) \rd\ell + \frac{1}{2} \int_{k \stepsize}^s \left( 3 \Exs
    \vecnorm{\discretized{r}_\ell - \discretized{r}_{k
        \stepsize}}{2}^2 + \Exs \vecnorm{\discretized{r}_{k
        \stepsize}}{2}^2 \right) \rd\ell \\& \quad + 2 (s - k \stepsize) \xi
    d / L.
  \end{align*}
Applying Gr\"{o}nwall's inequality yields
  \begin{align*}
    \Exs \vecnorm{\tilde{r}_s - \tilde{r}_{k \stepsize}}{2}^2 &\leq
    (e^{2 (s - k \stepsize)} - 1)\left( \sup_{k \stepsize \leq \ell
      \leq (k + 1) \stepsize} \Exs \vecnorm{\discretized{p}_\ell}{2}^2
    + \Exs \vecnorm{\discretized{r}_{k \stepsize}}{2}^2 + 2 \xi d / L
    \right)\\
    &\leq 14 \left(\Exs \vecnorm{\discretized{\theta}_{k \stepsize} -
      \thetastar}{2}^2 + \Exs \vecnorm{\discretized{p}_{k \stepsize}
    }{2}^2 + \Exs \vecnorm{\discretized{r}_{k \stepsize}}{2}^2 + d /
    L \right).
  \end{align*}
  In the second inequality, we plug in the bound for $\Exs
  \vecnorm{\discretized{p}_\ell}{2}^2$. Consequently,
  \begin{align*}
    \sup_{k \stepsize \leq s \leq (k + 1) \stepsize}\Exs
    \vecnorm{\tilde{r}_s}{2}^2 \leq 30 \left(\Exs
    \vecnorm{\discretized{\theta}_{k \stepsize} - \thetastar}{2}^2 +
    \Exs \vecnorm{\discretized{p}_{k \stepsize} }{2}^2 + \Exs
    \vecnorm{\discretized{r}_{k \stepsize}}{2}^2 + d / L \right),
  \end{align*}
  which completes the proof.


\subsubsection{Proof of Lemma~\ref{lemma:high-moments}}

For notational convenience, we assume $\thetastar = 0$ in the proof of
this lemma.  This assumption can be made without loss of generality.
Conditional on $x^{(k)}$, we have by the definition in
Algorithm~\ref{alg:main}: (we omit the superscripts and denote
$(\theta^{(k)}, p^{(k)}, r^{(k)})$ by $(\theta, p, r)$.)
\begin{align*}
  \mu (x^{(k)}) - x^{(k)} = & \left(
  \begin{array}{l}
    - \frac{\stepsize}{2L} \displaystyle\int_0^\stepsize \hat{g}_{t +
      k \stepsize} \left(\theta, \theta + \stepsize p\right) dt +
    \mu_{12} p +
    \mu_{13} {r} \\ - \frac{1}{L}
    \displaystyle\int_0^\stepsize \hat{g}_{t + k \stepsize}
    \left(\theta, \theta + \stepsize p\right) dt + (\mu_{22} - 1) {p} + \mu_{23} {r} \\ \frac{\mu_{31}}{L}
    \displaystyle\int_0^\stepsize \hat{g}_{t + k \stepsize}
    \left(\theta, \theta + \stepsize p\right) dt + \mu_{32} {p} +
    (\mu_{33} - 1) {r},
  \end{array}
  \right)
\end{align*}
    
Since $\hat{g}$ belongs to the convex hull of the curve $\nabla
U(\theta + t p)$, as assumed in the statement of the lemma, we have:
\begin{align*}
  \vecnorm{\frac{1}{ \stepsize} \int_0^\stepsize \hat{g}_{t + k
      \stepsize} \left(\theta, \theta + \stepsize p\right) dt - \nabla
    U (\theta)}{2} \leq & \sup_{t \in [0, \stepsize]} \vecnorm{\nabla
    U(\theta + t p) - \nabla U (\theta)}{2} \leq L \stepsize
  \vecnorm{p}{2},
\end{align*}
and using the smoothness of $U$, we can easily see that
\begin{align*}
  \vecnorm{\nabla U (\theta)}{2} \leq L \vecnorm{\theta -
    \thetastar}{2}, \forall \theta \in \real^d.
\end{align*}
    
Collecting the main terms and bound the rest of terms directly using
the norm of $x^{(k)}$, we obtain:
\begin{align*}
  \mu (x^{(k)}) - x^{(k)} &= \stepsize \left(
  \begin{array}{l}
    p^{(k)}\\ -\frac{1}{L} \nabla U (\theta^{(k)}) + \gamma
    r^{(k)}\\ - \gamma p^{(k)} - \xi r^{(k)}
  \end{array}
  \right) + \zeta_k\\
  &= \stepsize \left(
  \begin{matrix}
    0& I& 0\\ - \frac{\nabla^2 U (x^{(k)})}{L} I & 0 & \gamma I\\ 0&
    - \gamma I & - \xi I
  \end{matrix}
  \right) x^{(k)} + \zeta_k = \stepsize M_k x^{(k)} + \zeta_k.
\end{align*}
By Appendix~\ref{appnd:discretization}, it is easy to see that $\vecnorm{\zeta_k}{2} \leq C \stepsize^2 \kappa^2
\vecnorm{x^{(k)}}{2}$.
    
Therefore, we have:
\begin{multline*}
  \vecnorm{\mu(x^{(k)})}{S}^2 = \left( (I + \stepsize M_k) x^{(k)} +
  \zeta_k \right)^T S \left( (I + \stepsize M_k) x^{(k)} + \zeta_k
  \right)\\ \leq \vecnorm{x^{(k)}}{S}^2 + 2 \stepsize (x^{(k)})^T M_k
  S x^{(k)} + \stepsize^2 \matsnorm{M_k}{op}^2 \matsnorm{S}{op}
  \vecnorm{x^{(k)}}{2}^2 \\ + 2 \vecnorm{ \zeta_k}{2} \left(1 +
  \stepsize \matsnorm{M_k}{op} \right) \vecnorm{x^{(k)}}{2} +
  \vecnorm{\zeta_k}{2}^2 \\ \leq \left(1 - \frac{\stepsize}{5 \kappa^2
    + 50} \right) \vecnorm{x^{(k)}}{S}^2 + 6 \stepsize^2 \kappa^5
  \vecnorm{x^{(k)}}{S}^2.
\end{multline*}
For $\stepsize < c \kappa^{- 7}$ with some universal constant $c > 0$,
we have:
\begin{align*}
  \vecnorm{\mu(x^{(k)})}{S}^2 \leq \left( 1 - \frac{\stepsize}{10
    \kappa^2 + 100} \right) \vecnorm{x^{(k )}}{S}^2.
\end{align*}
Now we turn to deal with the stochastic part. By our construction, it
is easy to verify that $\matsnorm{\Sigma}{op} \leq C \kappa^2
\stepsize$ for some universal constant $C > 0$.
    
Letting $\xi_k \sim \mathcal{N}(0, \frac{1}{L} I)$ be independent from
$x^{(k)}$, we have:
\begin{align*}
  \Exs \vecnorm{x^{(k + 1)}}{S}^{2 \alpha} &= \Exs \vecnorm{
    S^{\frac{1}{2}}\mu (x^{(k)}) + (S \Sigma)^{\frac{1}{2}}
    \xi_k}{2}^{2 \alpha}\\ &\leq \sum_{j = 0}^{2 \alpha} \binom{2
    \alpha}{j} \Exs \vecnorm{S^{\frac{1}{2}} \mu (x^{(k)})}{2}^{j}
  \Exs \vecnorm{(S \Sigma)^{\frac{1}{2}} \xi_k}{2}^{2 \alpha -
    j}\\ &\leq \sum_{j = 0}^{2 \alpha} \binom{2 \alpha}{j} \left( \Exs
  \vecnorm{S^{\frac{1}{2}} \mu (x^{(k)})}{2}^{2 \alpha}
  \right)^{\frac{j}{2 \alpha}} \left( \Exs \vecnorm{(S
    \Sigma)^{\frac{1}{2}} \xi_k}{2}^{2 \alpha}\right)^{1 - \frac{j}{2
      \alpha}}\\ &= \left( \left(\Exs \vecnorm{\mu(x^{(k)})}{S}^{2
    \alpha}\right)^{\frac{1}{2 \alpha}} + \left( \Exs \vecnorm{(S
    \Sigma)^{\frac{1}{2}} \xi_k}{2}^{2 \alpha}\right)^{\frac{1}{2
      \alpha}} \right)^{2 \alpha}.
    \end{align*}
So we obtain:
\begin{align*}
  \left( \Exs \vecnorm{x^{(k + 1)}}{S}^{2 \alpha}
  \right)^{\frac{1}{2 \alpha}} \leq& \left(\Exs
  \vecnorm{\mu(x^{(k)})}{S}^{2 \alpha}\right)^{\frac{1}{2
      \alpha}} + \left( \Exs \vecnorm{(S \Sigma)^{\frac{1}{2}}
    \xi_k}{2}^{2 \alpha}\right)^{\frac{1}{2 \alpha}} \\ \leq&
  \left( 1 - \frac{\stepsize}{ 10 \kappa^2 + 100} \right)
  \left(\Exs \vecnorm{x^{(k)}}{S}^{2 \alpha}\right)^{\frac{1}{2
      \alpha}} + C \kappa \sqrt{\frac{\stepsize}{L} (d + 2
    \alpha)}.
\end{align*}
Noting that $\vecnorm{x^{(0)}}{2} \leq \frac{1}{L}$, by solving the recursion inequalities, we obtain:
\begin{align*}
  \left(\Exs \vecnorm{x^{(k)}}{2}^{2 \alpha}\right)^{\frac{1}{2
      \alpha}} \leq C \kappa^3 \sqrt{\frac{d + 2 \alpha}{L}}.
\end{align*}


\section{Auxiliary results for Lemmas~\ref{lemma:eig} and~\ref{lemma:eig_S}}
\label{app:discrete-time-rate}

This appendix is dedicated to proofs of two auxiliary
results---namely, Lemmas~\ref{LemFact3} and~\ref{lemma:f_bounds}---that
underlie the proofs of Lemmas~\ref{lemma:eig} and~\ref{lemma:eig_S}.


\subsection{Definitions of the functions $g_1$, $g_2$ and $g_3$}
\label{appnd:g}

The functions $g_1$, $g_2$ and $g_3$ are given by:
\begin{subequations}
\begin{align}
\label{EqnDefnGone}
g_1(\kappa) & = 4 + \tfrac{20}{\kappa^2} + \tfrac{56}{\kappa^3} + \tfrac{40}{\kappa^4} +
\tfrac{8}{\kappa^5} + \tfrac{20}{\kappa^6} + \tfrac{12}{\kappa^7} + \tfrac{1}{\kappa^8} +
\tfrac{2}{\kappa^9} + \tfrac{1}{\kappa^{10}} \\
\label{EqnDefnGtwo}
g_2(\kappa) & = \tfrac{1}{\kappa^3} + \tfrac{5}{\kappa^5} + \tfrac{11}{\kappa^6} + \tfrac{5}{\kappa^7} +
\tfrac{1}{\kappa^8} + \tfrac{2}{\kappa^9} + \tfrac{1}{\kappa^{10}}, \quad \mbox{and} \\
\label{EqnDefnGthree}
g_3(\kappa) & = 8 + \tfrac{24}{\kappa^2} + \tfrac{60}{\kappa^3} + \tfrac{41}{\kappa^4} +
\tfrac{10}{\kappa^5} + \tfrac{21}{\kappa^6} + \tfrac{12}{\kappa^7} + \tfrac{1}{\kappa^8} + \tfrac{2}{\kappa^9} +
\tfrac{1}{\kappa^{10}}.
\end{align}
\end{subequations}


\subsection{Proof of Lemma~\ref{LemFact3}}
\label{appnd:fact3_proof}
  
By inspecting the definition of $g_1$ in equation~\eqref{EqnDefnGone},
we see that $\forall\kappa\geq1$, $g_1(\kappa)$ always lies in the interval $[4,+\infty)$, which
implies the inequality $2 - \sqrt{ g_1(\kappa) } \leq 0$.

As for the second inequality, we first group the $\sqrt{g_1(\kappa)}$
term and rewrite inequality~\eqref{eq:fact3_2} into the following equivalent form:
\begin{align*}
2 \kappa + \frac{6}{5} \geq (\kappa-1) \sqrt{g_1(\kappa)}.
\end{align*}
Since both the left and right sides are non-negative, we can square
both sides, thereby obtaining
\begin{align*}
\left(2\kappa + \tfrac{6}{5} \right)^2 - (\kappa-1)^2 g_1(\kappa) \geq
0.
\end{align*}
Expanding the left hand side, we obtain the equivalent form of
equation~\eqref{eq:fact3_2}:
\begin{align*}
\tfrac{64}{5} \kappa - \tfrac{564}{25} - \tfrac{16}{\kappa} +
\tfrac{52}{\kappa^2} + \tfrac{16}{\kappa^3} -\tfrac{44}{\kappa^4} +
\tfrac{20}{\kappa^5} + \tfrac{3}{\kappa^6} - \tfrac{12}{\kappa^7} +
\tfrac{2}{\kappa^8} - \tfrac{1}{\kappa^{10}} \geq 0.
\end{align*}
Since $\kappa \geq 1$, we can divide by $\kappa$ on both sides and
obtain an equivalent inequality with a polynomial function of
$1/\kappa$:
\begin{align*}
g_6(1/\kappa) = \tfrac{64}{5} -\tfrac{564}{25\kappa} -
\tfrac{16}{\kappa^2} + \tfrac{52}{\kappa^3} + \tfrac{16}{\kappa^4} -
\tfrac{44}{\kappa^5} + \tfrac{20}{\kappa^6} + \tfrac{3}{\kappa^7} -
\tfrac{12}{\kappa^8} + \tfrac{2}{\kappa^9} - \tfrac{1}{\kappa^{11}} \geq
0.
\end{align*}
We can rewrite $g_6(1/\kappa)$ as a polynomial of
$\left(\frac{1}{\kappa} -\frac{1}{2}\right)$:

\begin{align*}
g_6(1/\kappa) & = \tfrac{201599}{51200} -\tfrac{49211}{25600}
\left(\tfrac{1}{\kappa} -\tfrac{1}{2}\right) + \tfrac{24025}{512}
\left(\tfrac{1}{\kappa} -\tfrac{1}{2}\right)^2 + \tfrac{2955}{256}
\left(\tfrac{1}{\kappa} -\tfrac{1}{2}\right)^3 \\
& - \tfrac{3397}{64} \left(\tfrac{1}{\kappa} -\tfrac{1}{2}\right)^4
-\tfrac{1399}{32} \left(\tfrac{1}{\kappa} -\tfrac{1}{2}\right)^5
-\tfrac{751}{16} \left(\tfrac{1}{\kappa} -\tfrac{1}{2}\right)^6
-\tfrac{381}{8} \left(\tfrac{1}{\kappa} -\tfrac{1}{2}\right)^7
\\
& - \tfrac{189}{8} \left(\tfrac{1}{\kappa} -\tfrac{1}{2}\right)^8
-\tfrac{47}{4} \left(\tfrac{1}{\kappa} -\tfrac{1}{2}\right)^9
-\tfrac{11}{2} \left(\tfrac{1}{\kappa} -\tfrac{1}{2}\right)^{10}
-\left(\tfrac{1}{\kappa} -\tfrac{1}{2}\right)^{11}.
\end{align*}
For all $\kappa$ such that $1/\kappa\in(0,1]$, we have
\begin{multline*}
g_6(1/\kappa) \geq \tfrac{201599}{51200} - \tfrac{49211}{25600}
\tfrac{1}{2} + \left(\tfrac{1}{\kappa} -\tfrac{1}{2}\right)^2
\Big(\tfrac{24025}{512} -\tfrac{2955}{256} \tfrac{1}{2} -
\tfrac{3397}{64} \left(\tfrac{1}{2}\right)^2 -\tfrac{1399}{32}
\left(\tfrac{1}{2}\right)^3 -\tfrac{751}{16}
\left(\tfrac{1}{2}\right)^4 \\
-\tfrac{381}{8} \left(\tfrac{1}{2}\right)^5 -\tfrac{189}{8}
\left(\tfrac{1}{2}\right)^6 -\tfrac{47}{4} \left(\tfrac{1}{2}\right)^7
-\frac{11}{2} \left(\tfrac{1}{2}\right)^{8}
-\left(\tfrac{1}{2}\right)^{9} \Big) \\ = \tfrac{38097}{12800} +
\left(\tfrac{1}{\kappa} -\tfrac{1}{2}\right)^2
\left(\tfrac{5523}{128}\right),
\end{multline*}
which is strictly positive.  This completes the proof of
Lemma~\ref{LemFact3}.


\subsection{Proof of Lemma~\ref{lemma:f_bounds}}
\label{appnd:lemma_f_bounds}

In order to prove the bounds in this lemma, we first examine the
monotonicity properties of the cubic function
\begin{align}
\label{EqnCubicFunction}
f(x) & \defn x^3 - g_2(\kappa) \cdot x^2 + \frac{1}{\kappa^9}
g_3(\kappa) \cdot x - \frac{1}{\kappa^{15}} g_3(\kappa).
\end{align}

\begin{lemma}
  \label{LemMonotonic}
  For any $\kappa \geq 1$, the function $f$ from
  equation~\eqref{EqnCubicFunction} has the following properties:
  \begin{enumerate}
  \item[(a)] It is monotonically increasing over the interval
    $(\frac{4}{\kappa^3} + \frac{40}{\kappa^5}, \infty)$.
  \item[(b)] It is monotonically increasing over the interval
    $(-\infty, \frac{4}{5\kappa^6})$.
  \end{enumerate}
\end{lemma}

Therefore, $\max\limits_{x \leq \frac{4}{5\kappa^6}} f(x) = f\left(
\frac{4}{5\kappa^6} \right)$ and $\min\limits_{x \geq
  \frac{4}{\kappa^3} + \frac{40}{\kappa^5}} f(x) = f\left(
\frac{4}{\kappa^3} + \frac{40}{\kappa^5} \right)$.  Then we simply
need to prove that $f\left( \frac{4}{5\kappa^6} \right)<0$ and
$f\left( \frac{4}{\kappa^3} + \frac{40}{\kappa^5} \right)>0$ to
obtain the result.

Now observe that
\begin{multline*}
f\left( \tfrac{4}{\kappa^3} + \tfrac{40}{\kappa^5} \right) =
\tfrac{48}{\kappa^9} + \tfrac{1520}{\kappa^{11}} -
\tfrac{144}{\kappa^{12}} + \tfrac{15920}{\kappa^{13}} -
\tfrac{3120}{\kappa^{14}} + \tfrac{54600}{\kappa^{15}} \\ -
\tfrac{16812}{\kappa^{16}} - \tfrac{6224}{\kappa^{17}} -
\tfrac{256}{\kappa^{18}} - \tfrac{2793}{\kappa^{19}} -
\tfrac{766}{\kappa^{20}} + \tfrac{467}{\kappa^{21}} +
\tfrac{32}{\kappa^{22}} + \tfrac{79}{\kappa^{23}} +
\tfrac{38}{\kappa^{24}} - \tfrac{1}{\kappa^{25}},
\end{multline*}
from which we can see that
\begin{align*}
  f\left( \tfrac{4}{\kappa^3} + \tfrac{40}{\kappa^5} \right) \geq
  \tfrac{48}{\kappa^9} + \tfrac{1376}{\kappa^{11}} +
  \tfrac{12800}{\kappa^{13}} + \tfrac{27749}{\kappa^{15}} +
  \tfrac{467}{\kappa^{21}} > 0,
\end{align*}
using the fact that $1/\kappa\in(0,1]$.  Moreover, we also have
\begin{align*}
f\left( \tfrac{4}{5\kappa^6} \right) = - \tfrac{56}{25\kappa^{15}} -
\tfrac{8}{\kappa^{17}}- \tfrac{2316}{125\kappa^{18}} -
\tfrac{57}{5\kappa^{19}} - \tfrac{66}{25\kappa^{20}} -
\tfrac{137}{25\kappa^{21}} - \tfrac{76}{25\kappa^{22}} -
\tfrac{1}{5\kappa^{23}} - \tfrac{2}{5\kappa^{24}}
-\tfrac{1}{5\kappa^{25}},
\end{align*}
which is negative.  Therefore, we conclude that for any $\kappa>1$,
the cubic function $f$ satisfies the inequalities
\begin{align*}
  f(x) < 0 \quad \mbox{if $x \leq \frac{4}{5\kappa^6}$, and} \qquad
  f(x) > 0 \quad \mbox{ if $x \geq \frac{4}{\kappa^3} +
    \frac{40}{\kappa^5}$.}
\end{align*}


\subsubsection{Proof of Lemma~\ref{LemMonotonic}}

We divide our proof into separate parts, corresponding to claims (a)
and (b) in the lemma statement.  For both parts, we establish
monotonicity of the function $f$ from
equation~\eqref{EqnCubicFunction} by studying its derivative
\begin{align}
  \label{EqnCubicDerivative}
  f'(x) = 3 x^2 - 2 g_2(\kappa) \cdot x + \frac{1}{\kappa^9} g_3(\kappa).
\end{align}

\paragraph{Proof of part (a):} 
By inspection, for large enough $x$, the quadratic function $f'$ is
positive, and hence the function $f$ is monotonically increasing in
this range.  Concretely, we claim that $f'$ remains positive for all
$x > \frac{4}{\kappa^3} + \frac{40}{\kappa^5}$.

Our strategy is to compute the solutions $x^*_{\pm}$ to the quadratic
equation $f'(x) = 0$, and prove that the larger one $x^*_{+}$
satisfies the lower bound
\begin{align}
  \label{EqnLargerBound}
  x^*_{+} \leq \frac{4}{\kappa^3} + \frac{40}{\kappa^5} \quad
  \mbox{for any $\kappa \geq 1$.}
\end{align}
In detail, the two solutions to the quadratic equation $f'(x) = 0$ are
given by the pair $x^*_{\pm} = \frac{1}{3} \left( g_4(\kappa) \pm
\sqrt{g_5(\kappa)} \right)$, where we define
\begin{subequations}
\begin{align}
    \label{eq:g_4}  
    g_4(\kappa) = \tfrac{1}{\kappa^3} + \tfrac{5}{\kappa^5} + \tfrac{11}{\kappa^6} + \tfrac{5}{\kappa^7} +
    \tfrac{1}{\kappa^8} + \tfrac{2}{\kappa^9} + \tfrac{1}{\kappa^{10}}
\end{align}
and
\begin{align}
    \label{eq:g_5}  
    g_5(\kappa) = \tfrac{1}{\kappa^6} + \tfrac{10}{\kappa^8} - \tfrac{2}{\kappa^9} +
    \tfrac{35}{\kappa^{10}} + \tfrac{40}{\kappa^{11}} - \tfrac{5}{\kappa^{12}} - \tfrac{1}{\kappa^{13}}
    + \tfrac{37}{\kappa^{14}} + \tfrac{1}{\kappa^{15}} + \tfrac{7}{\kappa^{16}} +
    \tfrac{11}{\kappa^{17}} + \tfrac{1}{\kappa^{19}} + \tfrac{1}{\kappa^{20}}.
\end{align}
\end{subequations}

From the fact that $\tfrac{1}{\kappa}\in(0,1]$, it follows that
\begin{subequations}
  \begin{align}
    \label{eq:appnd_g4}
g_4(\kappa) \leq \frac{1}{\kappa^3} + \frac{25}{\kappa^5}, \quad \mbox{and} \quad
g_5(\kappa) \leq \frac{11}{\kappa^6} + \frac{125}{\kappa^{10}}.
\end{align}
Hence
\begin{align}
\sqrt{g_5(\kappa)} \leq \sqrt{\frac{11}{\kappa^6}} + \sqrt{\frac{125}{\kappa^{10}}}
\leq \frac{4}{\kappa^3} + \frac{12}{\kappa^5}.
\label{eq:appnd_g5}
\end{align}
\end{subequations}
Combining equations~\eqref{eq:appnd_g4} and~\eqref{eq:appnd_g5} yields
the bound~\eqref{EqnLargerBound}, and hence completes the proof of
part (a).


\paragraph{Proof of part (b):}

Recall the solutions to the quadratic equation $f'(x) = 0$ that we
computed in the previous section.  For this part, it suffices to show
that the smaller solution $x^*_{-} = \frac{1}{3} \left( g_4(\kappa) -
\sqrt{g_5(\kappa)} \right)$ satisfies the bound
\begin{align}
  \label{EqnSmallerBound}
x^*_{-} & \geq \frac{4}{5\kappa^6} \quad \mbox{for any $\kappa \geq
  1$.}
\end{align}
We begin by noting that
\begin{align}
\frac{1}{3} \left( g_4(\kappa) - \sqrt{g_5(\kappa)} \right) &=
\frac{1}{3} \frac{g_4(\kappa)^2 - g_5(\kappa)}{g_4(\kappa) +
  \sqrt{g_5(\kappa)}} \nonumber \\
& = \frac{1}{\kappa^9} \frac{ \left( 8 + \frac{24}{\kappa^2} +
  \frac{60}{\kappa^3} + \frac{41}{\kappa^4} +\frac{10}{\kappa^5} +
  \frac{21}{\kappa^6} + \frac{12}{\kappa^7} + \frac{1}{\kappa^8} +
  \frac{2}{\kappa^9} + \frac{1}{\kappa^{10}} \right)}{g_4(\kappa) +
  \sqrt{g_5(\kappa)}}.
  \label{eq:appnd_fact2_1}
\end{align}
Equation~\eqref{eq:appnd_g5} states that $\sqrt{g_5(\kappa)} \leq
\tfrac{4}{\kappa^3} + \tfrac{12}{\kappa^5}$, and hence for any $1/\kappa\in(0,1]$,
\begin{align}
  g_4(\kappa) + \sqrt{g_5(\kappa)}
  & \leq \tfrac{5}{\kappa^3} +
  \tfrac{17}{\kappa^5} + \tfrac{11}{\kappa^6} + \tfrac{5}{\kappa^7} + \tfrac{1}{\kappa^8} +
  \tfrac{2}{\kappa^9} + \tfrac{1}{\kappa^{10}} \nonumber\\ 
  &\leq \tfrac{5}{\kappa^3} +
  \tfrac{17}{\kappa^5} + \tfrac{20}{\kappa^6} \nonumber \\
\label{eq:appnd_fact2_2}
& \leq
\frac{ 8 + \tfrac{24}{\kappa^2} +
\tfrac{60}{\kappa^3} + \tfrac{41}{\kappa^4} + \tfrac{10}{\kappa^5} +
\tfrac{21}{\kappa^6} + \tfrac{12}{\kappa^7} + \tfrac{1}{\kappa^8} +
\tfrac{2}{\kappa^9} + \tfrac{1}{\kappa^{10}} }{\kappa^3}.
\end{align}
Since $g_4(\kappa) + \sqrt{g_5(\kappa)}$ is positive for $\kappa \geq
1$, we can substitute the bound~\eqref{eq:appnd_fact2_2} into
equation~\eqref{eq:appnd_fact2_1}, thereby finding that
\begin{align*}
x^*_{-} &= \frac{1}{3} \left( g_4(\kappa) - \sqrt{g_5(\kappa)} \right)
\\ &= \frac{1}{\kappa^9} \frac{  8 + \tfrac{24}{\kappa^2} +
\tfrac{60}{\kappa^3} + \tfrac{41}{\kappa^4} + \tfrac{10}{\kappa^5} +
\tfrac{21}{\kappa^6} + \tfrac{12}{\kappa^7} + \tfrac{1}{\kappa^8} +
\tfrac{2}{\kappa^9} + \tfrac{1}{\kappa^{10}} }{g_4(\kappa) +
  \sqrt{g_5(\kappa)}} \\ 
  &\geq \frac{1}{\kappa^6} \geq \frac{4}{5\kappa^6},
\end{align*}
which completes the proof of the bound~\eqref{EqnSmallerBound}.



\section{Details of Algorithm~\ref{alg:main}}
\label{appnd:discretization}

This section is devoted to explicit definition of all constants
involved in Algorithm~\ref{alg:main}, as well as a derivation of how
the algorithm's updates follows from integrating
equations~\eqref{eq:underdamp-euler}--\eqref{eq:underdamp-correction-general}.
We begin by defining the constants precisely:
\begin{align*}
\mu_{12} &= \left( 1 + \frac{\gamma^2}{\xi^2} \right) \stepsize
    - \frac{\gamma^2}{2\xi} \stepsize^2
    - \frac{\gamma^2}{\xi^3} \left( 1 - e^{-\xi\stepsize} \right) \\
\mu_{13} &= \frac{\gamma}{\xi} \stepsize +
    \frac{\gamma}{\xi^2} \left( e^{-\xi\stepsize} - 1 \right) \\
\mu_{22} &= 1 + \frac{\gamma^2}{\xi^2} \left( 1 - \xi\stepsize - e^{-\xi\stepsize} \right) \\
\mu_{23} &= \frac{\gamma}{\xi} \left( 1 - e^{-\xi\stepsize} \right) \\
\mu_{31} &= \frac{\gamma}{\xi} - \frac{\gamma}{\xi^2}
    \frac{1-e^{-\xi\stepsize}}{\stepsize} \\
\mu_{32} &= \frac{\gamma^3}{\xi^2}\stepsize 
    + \frac{\gamma^3}{\xi^2} \stepsize e^{-\xi\stepsize}
    - \left( \frac{2\gamma^3}{\xi^3} + \frac{\gamma}{\xi}
    \right) \left( 1 - e^{-\xi\stepsize} \right) \\
\mu_{33} &= e^{-\xi\stepsize} +\frac{\gamma^2}{\xi} \stepsize e^{-\xi\stepsize}
    - \frac{\gamma^2}{\xi^2} \left( 1 - e^{-\xi\stepsize} \right),
\end{align*}
as well as
\begin{align*}
\sigma_{11} & = \frac{2\gamma^2}{\xi^3} \stepsize 
- \frac{2\gamma^2}{\xi^2} \stepsize^2
+ \frac{2\gamma^2}{3\xi} \stepsize^3
- \frac{4 \gamma^2}{\xi^3} \stepsize e^{-\stepsize  \xi }
+ \frac{\gamma^2}{\xi^4} \left( 1 - e^{-2 \xi\stepsize} \right) \\
\sigma_{12} & = \frac{\gamma^2}{\xi^3 L}
\left( \xi\stepsize - \left( 1 - e^{-\xi\stepsize} \right) \right)^2 \\
\sigma_{22} & = \frac{2 \gamma^2}{\xi} \stepsize
- \frac{4 \gamma^2}{\xi^2} \left( 1 - e^{-\xi\stepsize} \right)
+ \frac{\gamma^2}{\xi^2} \left( 1 - e^{-2\xi\stepsize} \right) \\
\sigma_{13} & = 
- \frac{ \gamma^3 }{\xi^2} \stepsize^2 \left( 2 e^{ - \xi \stepsize } + 1 \right)
+ \left(\frac{2\gamma^3}{\xi^3}
- \frac{\gamma^3}{\xi^3} e^{-2 \xi \stepsize}
- \frac{4\gamma^3}{\xi^3} e^{-\stepsize  \xi }
- \frac{ 2\gamma }{ \xi } e^{-\stepsize \xi } \right) \stepsize  
+ \left( \frac{3\gamma^3}{2 \xi^4} + \frac{\gamma}{\xi^2} \right)
\left(1 - e^{-2 \xi \stepsize}\right) \\
\sigma_{23} & = 
\frac{\gamma^3}{\xi^2} \left(e^{-2 \xi \stepsize} - 2 e^{-\xi \stepsize} - 2\right) \stepsize 
+ \frac{3\gamma^3}{2\xi^3} \left(e^{-2\xi\stepsize} - 4e^{-\xi\stepsize} + 3\right)
+ \frac{\gamma}{\xi} \left(1 - e^{-\xi \stepsize} \right)^2 \\
\sigma_{33} & = 
-\frac{\gamma^4}{\xi^2} \stepsize^2 e^{-2 \xi \stepsize} 
+ \left( -\frac{2\gamma^2}{\xi} e^{-2 \xi \stepsize}
+ \frac{\gamma^4}{\xi^3} \left( - 3 e^{-2 \xi \stepsize} + 4 e^{- \xi \stepsize} + 2 \right)
\right) \stepsize \\
&+ \frac{\gamma^4}{2 \xi^4} \left(-5e^{-2 \eta  \xi } + 16 e^{-\eta  \xi }-11\right)
+ \frac{\gamma ^2}{\xi^2} \left(-3e^{-2 \eta  \xi } + 4 e^{-\eta  \xi }-1\right)
+ \left(1-e^{-2 \eta  \xi }\right).
\end{align*}
\noindent Given these definitions, we now demonstrate how to obtain
Algorithm~\ref{alg:main} via integrating
equations~\eqref{eq:underdamp-euler} through
to~\eqref{eq:underdamp-correction-general}.

\paragraph{First step:}
The first step involving the Ornstein-Uhlenbeck process can be explicitly solved as (let $B_0 = 0$):
\begin{align}
  \label{eq:out_underdamp-euler}
\begin{cases}
    \hat{\theta}_{t} = \discretized{\theta}_{k \stepsize} +
    (t-k\stepsize) \discretized{p}_{k \stepsize},\\ 
    \hat{r}_t = e^{ - \xi (t-k\stepsize) } \discretized{r}_{k \stepsize} 
    - \frac{\gamma}{\xi} \left( 1 - e^{-\xi(t-k\stepsize)} \right) \discretized{p}_{k \stepsize} 
    + \sqrt{\frac{2\xi}{L}}
    \displaystyle\int_{k\stepsize}^t e^{-\xi(t-s)} \rd B_{s}^r.
\end{cases}
\end{align}
It is worth noting that we have set at the beginning of every step $\hat{\theta}_{k \stepsize} = \discretized{\theta}_{k \stepsize}$ and $\hat{r}_{k \stepsize} = \discretized{r}_{k \stepsize}$.
 
\paragraph{Second step:}  Next we integrate the right hand side of
equation~\eqref{eq:integration-p} to obtain $\discretized{p}$:
\begin{align}
\hat{p}_t = \discretized{p}_{k\stepsize} - \frac{1}{L\stepsize}
\int_{k\stepsize}^{t} \left(\displaystyle\int_{k \stepsize}^{(k + 1)
  \stepsize} \hat{g}_s \rd s \right) \rd \varsigma + \gamma
\int_{k\stepsize}^{t} \hat{r}_{s} \rd s
\end{align}
If function $U$ takes the form of $U(\theta) = \sum_i u_i(y_i^\rT
\theta)$, then we directly take $\hat{g}_s=\nabla U( \hat{\theta}_s )$
and calculate the integral:
\begin{align*}
\int_{k \stepsize}^{(k + 1) \stepsize} \hat{g}_s \rd s &=
\int_{k\stepsize}^{(k + 1) \stepsize} \sum_i u'_i\left( y_i^\rT \left(
\discretized{\theta}_{k \stepsize} + (s-k\stepsize) \discretized{p}_{k
  \stepsize} \right) \right) y_i \ \rd s \\ &= \sum_i
\int_{k\stepsize}^{(k + 1) \stepsize} u'_i\left( y_i^\rT \left(
\discretized{\theta}_{k \stepsize} + (s-k\stepsize) \discretized{p}_{k
  \stepsize} \right) \right) \rd s \ y_i.
\end{align*}
Using the Newton-Leibniz formula, we obtain that
\begin{align*}
\int_{k\stepsize}^{(k + 1) \stepsize} \hat{g}_s \rd s = \sum_i \left(
u_i\left( y_i^\rT ( \discretized{\theta}_{k \stepsize} + \stepsize
\discretized{p}_{k \stepsize} ) \right) - u_i\left( y_i^\rT
\discretized{\theta}_{k \stepsize} \right) \right) \frac{y_i}{y_i^\rT
  \discretized{p}_{k \stepsize}}.
\end{align*}
Hence
\begin{align*}
    \int_{k\stepsize}^{t} \left(\displaystyle\int_{k \stepsize}^{(k +
      1) \stepsize} \hat{g}_s \rd s \right) \rd \varsigma &=
    (t-k\stepsize) \int_{k\stepsize}^{(k + 1) \stepsize} \hat{g}_s \rd
    s \\
    & = (t-k\stepsize) \sum_i \left( u_i\left( y_i^\rT (
    \discretized{\theta}_{k \stepsize} + \stepsize \discretized{p}_{k
      \stepsize} ) \right) - u_i\left( y_i^\rT \discretized{\theta}_{k
      \stepsize} \right) \right) \frac{y_i}{y_i^\rT \discretized{p}_{k
        \stepsize}}.
\end{align*}

For $\displaystyle \int_{k\stepsize}^{t} \hat{r}_s \rd s$, we use
Fubini's theorem and obtain that
\begin{align*}
\int_{k\stepsize}^{t} \hat{r}_s \rd s 
&= \int_{k\stepsize}^{t} \left(
e^{ - \xi (s-k\stepsize) } \discretized{r}_{k \stepsize} 
- \frac{\gamma}{\xi} \left( 1 - e^{-\xi(s-k\stepsize)} \right) \discretized{p}_{k \stepsize} 
+ \sqrt{\frac{2\xi}{L}} \displaystyle\int_{k\stepsize}^s e^{-\xi(s-\hat{s})} \rd B_{\hat{s}}^r
\right) \rd s \\ 
&= \frac{1}{\xi} \left( 1 - e^{-\xi\discStep} \right) \discretized{r}_{k \stepsize} 
+ \frac{\gamma}{\xi^2} \left( 1 - \xi\discStep - e^{-\xi\discStep} \right) \discretized{p}_{k \stepsize} \\
&+ \sqrt{\frac{2\xi}{L}} \int_{k\stepsize}^{t} 
\left( \int_{k\stepsize}^{s} e^{-\xi(s-\hat{s})} \rd B_{\hat{s}}^r \right) \rd s \\ 
&= \frac{\gamma}{\xi^2} \left( 1 - \xi\discStep - e^{-\xi\discStep} \right) \discretized{p}_{k \stepsize}
+ \frac{1}{\xi} \left( 1 - e^{-\xi\discStep} \right) \discretized{r}_{k \stepsize} \\
&+ \sqrt{\frac{2\xi}{L}} \int_{k\stepsize}^{t} \left( \int_{\hat{s}}^{t} e^{-\xi(s-\hat{s})} \rd s
\right) \rd B_{\hat{s}}^r \\ 
&= \frac{\gamma}{\xi^2} \left( 1 - \xi\discStep - e^{-\xi\discStep} \right) \discretized{p}_{k \stepsize}
+ \frac{1}{\xi} \left( 1 - e^{-\xi\discStep} \right) \discretized{r}_{k \stepsize} \\
&+ \sqrt{\frac{2}{\xi L}} \int_{k\stepsize}^{t} \left( 1 - e^{-\xi(t-\hat{s})} \right) \rd B_{\hat{s}}^r.
\end{align*}
Therefore, we obtain $\hat{p}_t$ from explicit integration:
\begin{align}
   \hat{p}_t & = \underbrace{ - \frac{t-k\stepsize}{L\stepsize}
     \sum_i \left( u_i\left( y_i^\rT ( \discretized{\theta}_{k
       \stepsize} + \stepsize \discretized{p}_{k \stepsize} ) \right)
     - u_i\left( y_i^\rT \discretized{\theta}_{k \stepsize} \right)
     \right) \frac{y_i}{y_i^\rT \discretized{p}_{k \stepsize}}
   }_{T_1(t)} \nonumber\\ 
   &+ \underbrace{ \left( 1 + \frac{\gamma^2}{\xi^2} \left( 1 - \xi\discStep - e^{-\xi\discStep} \right) \right) \discretized{p}_{k \stepsize}
     + \frac{\gamma}{\xi} \left( 1 - e^{-\xi\discStep} \right) \discretized{r}_{k \stepsize}
   }_{T_2(t)} \nonumber\\
   &+ \underbrace{ \sqrt{\frac{2\gamma^2}{\xi L}} \int_{k\stepsize}^{t} \left( 1 - e^{-\xi(t-\hat{s})} \right) \rd B_{\hat{s}}^r
   }_{T_3(t)}. \label{eq:discretized_pt}
\end{align}

\paragraph{Third step:}
Next observe that
\begin{align*}
  \discretized{\theta}_{(k+1)\stepsize} &=
  \discretized{\theta}_{k\stepsize} +
  \int_{k\stepsize}^{(k+1)\stepsize} \hat{p}_t \rd t \; = \;
  \discretized{\theta}_{k\stepsize} +
  \int_{k\stepsize}^{(k+1)\stepsize} \left( T_1(t) + T_2(t) + T_3(t)
  \right) \rd t,
\end{align*}
where
\begin{align*}
\int_{k\stepsize}^{(k+1)\stepsize} T_1(t) \rd t &= -
\frac{1}{L\stepsize} \int_{k\stepsize}^{(k+1)\stepsize}
(t-k\stepsize) \rd t \cdot \int_{k\stepsize}^{(k + 1) \stepsize}
\hat{g}_s \rd s \\ &= - \frac{\stepsize}{2 L} \sum_i \left( u_i\left(
y_i^\rT ( \discretized{\theta}_{k \stepsize} + \stepsize
\discretized{p}_{k \stepsize} ) \right) - u_i\left( y_i^\rT
\discretized{\theta}_{k \stepsize} \right) \right) \frac{y_i}{y_i^\rT
  \discretized{p}_{k \stepsize}},
\end{align*}
\begin{align*}
    \int_{k\stepsize}^{(k+1)\stepsize} T_2(t) \rd t 
    = \left( \left( 1 + \frac{\gamma^2}{\xi^2} \right) \stepsize
    - \frac{\gamma^2}{2\xi} \stepsize^2
    - \frac{\gamma^2}{\xi^3} \left( 1 - e^{-\xi\stepsize} \right) \right) \discretized{p}_{k\stepsize} 
    + \left( \frac{\gamma}{\xi} \stepsize +
    \frac{\gamma}{\xi^2} \left( e^{-\xi\stepsize} - 1 \right) \right) \discretized{r}_{k\stepsize},
\end{align*}
and
\begin{align*}
  \int_{k\stepsize}^{(k+1)\stepsize} T_3(t) \rd t 
  &= \sqrt{\frac{2\gamma^2}{\xi L}} \int_{k\stepsize}^{(k+1)\stepsize} \left( \int_{k\stepsize}^{t} \left( 1 - e^{-\xi(t-s)} \right) \rd B_{s}^r \right) \rd t \\
  & = \sqrt{\frac{2\gamma^2}{\xi L}} \int_{k\stepsize}^{(k+1)\stepsize}
  \left( \int_{s}^{(k+1)\stepsize} \left( 1 - e^{-\xi(t-s)} \right) \rd t \right)
  \rd B_{s}^r \\
  &= \sqrt{\frac{2\gamma^2}{\xi L}} \int_{k\stepsize}^{(k+1)\stepsize}
  \left( \left( (k+1)\stepsize- s \right) + \frac{1}{\xi} \left( e^{-\xi\left( (k+1)\stepsize- s \right)} - 1 \right) \right) \rd B_{s}^r.
\end{align*}
Putting together the pieces, we find that
\begin{multline*}
    \discretized{\theta}_{(k+1)\stepsize} =
    \discretized{\theta}_{k\stepsize} - \frac{\stepsize}{2 L} \sum_i
    \left( u_i\left( y_i^\rT ( \discretized{\theta}_{k \stepsize} +
    \stepsize \discretized{p}_{k \stepsize} ) \right) - u_i\left(
    y_i^\rT \discretized{\theta}_{k \stepsize} \right) \right)
    \frac{y_i}{y_i^\rT \discretized{p}_{k \stepsize}} \\
+ \left( \left( 1 + \frac{\gamma^2}{\xi^2} \right) \stepsize
    - \frac{\gamma^2}{2\xi} \stepsize^2
    - \frac{\gamma^2}{\xi^3} \left( 1 - e^{-\xi\stepsize} \right) \right) \discretized{p}_{k\stepsize} 
    + \left( \frac{\gamma}{\xi} \stepsize +
    \frac{\gamma}{\xi^2} \left( e^{-\xi\stepsize} - 1 \right) \right) \discretized{r}_{k\stepsize}
\\
+ \sqrt{\frac{2\gamma^2}{\xi L}} \int_{k\stepsize}^{(k+1)\stepsize}
  \left( \left( (k+1)\stepsize- s \right) + \frac{1}{\xi} \left( e^{-\xi\left( (k+1)\stepsize- s \right)} - 1 \right) \right) \rd B_{s}^r.
\end{multline*}
We then calculate $\discretized{r}_{(k+1)\stepsize}$:
\begin{align*}
    e^{\xi \stepsize} \discretized{r}_{(k+1)\stepsize} -
    \discretized{r}_{k\stepsize} 
    &= \int_{k\stepsize}^{(k+1)\stepsize}
    \rd \left( e^{\xi (t-k\stepsize)} \discretized{r}_t \right) \\ 
    &=
    \int_{k\stepsize}^{(k+1)\stepsize} e^{\xi(t-k\stepsize)} \left( -
    \gamma \hat{p}_t \rd t + \sqrt{2 \xi/ L}\ \rd B_t^r
    \right) \\ 
    &= - \gamma \int_{k\stepsize}^{(k+1)\stepsize}
    e^{\xi(t-k\stepsize)} \hat{p}_t \rd t + \sqrt{\frac{2 \xi}{L}}
    \int_{k\stepsize}^{(k+1)\stepsize} e^{\xi(t-k\stepsize)} \ \rd B_t^r \\ 
    &= - \gamma \int_{k\stepsize}^{(k+1)\stepsize}
    e^{\xi(t-k\stepsize)} \left( T_1(t) + T_2(t) + T_3(t) \right) \rd t
    + \sqrt{\frac{2 \xi}{L}} \int_{k\stepsize}^{(k+1)\stepsize}
    e^{\xi(t-k\stepsize)} \ \rd B_t^r.
\end{align*}
Term $- \gamma \int_{k\stepsize}^{(k+1)\stepsize} e^{\xi(t-k\stepsize)} T_1(t) \rd t$ equals to:
\begin{align*}
    \lefteqn{- \gamma \int_{k\stepsize}^{(k+1)\stepsize}
      e^{\xi(t-k\stepsize)} T_1(t) \rd t} \\ &=
    \frac{\gamma}{L\stepsize} \int_{k\stepsize}^{(k+1)\stepsize}
    (t-k\stepsize) e^{\xi(t-k\stepsize)} \rd t \cdot
    \int_{k\stepsize}^{(k + 1) \stepsize} \hat{g}_s \rd s \\ &=
    e^{\xi\stepsize} \left( \frac{\gamma}{L\xi} -
    \frac{\gamma}{L\xi^2} \frac{1-e^{-\xi\stepsize}}{\stepsize}
    \right) \cdot \sum_i \left( u_i\left( y_i^\rT (
    \discretized{\theta}_{k \stepsize} + \stepsize \discretized{p}_{k
      \stepsize} ) \right) - u_i\left( y_i^\rT \discretized{\theta}_{k
      \stepsize} \right) \right) \frac{y_i}{y_i^\rT \discretized{p}_{k
        \stepsize}}.
\end{align*}
Term $- \gamma \int_{k\stepsize}^{(k+1)\stepsize}
e^{\xi(t-k\stepsize)} T_2(t) \rd t$ equals to:
\begin{align*}
    \lefteqn{- \gamma \int_{k\stepsize}^{(k+1)\stepsize}
    e^{\xi(t-k\stepsize)} T_2(t) \rd t} \\ 
    &= e^{\xi\stepsize}
    \left( \frac{\gamma^3}{\xi^2}\stepsize 
    + \frac{\gamma^3}{\xi^2} \stepsize e^{-\xi\stepsize}
    - \left( \frac{2\gamma^3}{\xi^3} + \frac{\gamma}{\xi}
    \right) \left( 1 - e^{-\xi\stepsize} \right) \right)
    \discretized{p}_{k \stepsize}  \\
    &+ e^{\xi\stepsize} \left(
    \frac{\gamma^2}{\xi} \stepsize e^{-\xi\stepsize}
    - \frac{\gamma^2}{\xi^2} \left( 1 - e^{-\xi\stepsize} \right)
    \right) \discretized{r}_{k \stepsize}.
\end{align*}
Term $- \gamma \int_{k\stepsize}^{(k+1)\stepsize}
e^{\xi(t-k\stepsize)} T_3(t) \rd t$ can be calculated to be:
\begin{align*}
    \lefteqn{- \gamma \int_{k\stepsize}^{(k+1)\stepsize}
    e^{\xi(t-k\stepsize)} T_3(t) \rd t} \\ 
    &= - \sqrt{\frac{2\gamma^4}{\xi L}} 
    \int_{k\stepsize}^{(k+1)\stepsize} \left(
    \int_{k\stepsize}^{t} e^{\xi(t-k\stepsize)} \left( 1 - e^{-\xi(t-s)} \right)
    \rd B_{s}^r \right) \rd t \\ 
    &= - \sqrt{\frac{2\gamma^4}{\xi L}} 
    \int_{k\stepsize}^{(k+1)\stepsize} \left(
    \int_{s}^{(k+1)\stepsize} \left( e^{\xi(t-k\stepsize)} - e^{\xi(s-k\stepsize)} \right) \rd t \right) \rd B_{s}^r \\ 
    &= \sqrt{\frac{2\gamma^4}{\xi L}} 
    \int_{k\stepsize}^{(k+1)\stepsize} \left(
    e^{\xi(s - k\stepsize)} \left( (k+1)\stepsize-s \right) 
    - \frac{1}{\xi}\left(e^{\xi\stepsize} -
    e^{\xi(s-k\stepsize)}\right) \right) \rd B_{s}^r.
\end{align*}
Summing the terms together, we obtain that
\begin{align*}
    \discretized{r}_{(k+1)\stepsize} 
    &= e^{-\xi \stepsize}
    \discretized{r}_{k\stepsize} \\ 
    &+ \left( \frac{\gamma}{L\xi} -
    \frac{\gamma}{L\xi^2} \frac{1-e^{-\xi\stepsize}}{\stepsize}
    \right) \cdot \sum_i \left( u_i\left( y_i^\rT (
    \discretized{\theta}_{k \stepsize} + \stepsize \discretized{p}_{k
      \stepsize} ) \right) - u_i\left( y_i^\rT \discretized{\theta}_{k
      \stepsize} \right) \right) \frac{y_i}{y_i^\rT \discretized{p}_{k
        \stepsize}} \\ 
    &+ \left( \frac{\gamma^3}{\xi^2}\stepsize 
    + \frac{\gamma^3}{\xi^2} \stepsize e^{-\xi\stepsize}
    - \left( \frac{2\gamma^3}{\xi^3} + \frac{\gamma}{\xi}
    \right) \left( 1 - e^{-\xi\stepsize} \right) \right)
    \discretized{p}_{k \stepsize}
    + \left(
    \frac{\gamma^2}{\xi} \stepsize e^{-\xi\stepsize}
    - \frac{\gamma^2}{\xi^2} \left( 1 - e^{-\xi\stepsize} \right)
    \right) \discretized{r}_{k \stepsize} \\ 
    &+ \sqrt{\frac{2 \xi}{L}}
    \int_{k\stepsize}^{(k+1)\stepsize} \left(
    \frac{\gamma^2}{\xi} \left((k+1)\stepsize-s\right) e^{ - \xi \left((k+1)\stepsize-s\right) }
    + \left( 1 + \frac{\gamma^2}{\xi^2} \right) e^{-\xi\left((k+1)\stepsize-s\right)}
    - \frac{\gamma^2}{\xi^2} 
    \right) \rd B_{s}^r.
\end{align*}
For $\discretized{p}_{(k+1)\stepsize}$, we directly know from
equation~\eqref{eq:discretized_pt} that
\begin{multline*}
\discretized{p}_{(k+1)\stepsize} 
   = - \frac{1}{L} \sum_i \left(
   u_i\left( y_i^\rT ( \discretized{\theta}_{k \stepsize} + \stepsize
   \discretized{p}_{k \stepsize} ) \right) - u_i\left( y_i^\rT
   \discretized{\theta}_{k \stepsize} \right) \right)
   \frac{y_i}{y_i^\rT \discretized{p}_{k \stepsize}} \\ 
   + \left( 1 + \frac{\gamma^2}{\xi^2} \left( 1 - \xi\stepsize - e^{-\xi\stepsize} \right) \right) \discretized{p}_{k\stepsize} 
   + \frac{\gamma}{\xi} \left( 1 - e^{-\xi\stepsize} \right) \discretized{r}_{k \stepsize} \\
   + \sqrt{\frac{2\gamma^2}{\xi L}} \int_{k\stepsize}^{(k+1)\stepsize} \left( 1 - e^{-\xi((k+1)\stepsize - s)} \right) \rd B_{s}^r.
\end{multline*}

Therefore, $\discretized{x}_{(k+1)} = (\discretized{\theta}_{(k+1)},
\discretized{p}_{(k+1)}, \discretized{r}_{(k+1)})$ conditioning on
$\discretized{x}_{(k)}$ follows a normal distribution.  We calculate
its mean and covariance below.  We first find
$\Ep{\discretized{x}_{(k+1)}}$ using the property of It\^o integral:

\begin{align*}
    \Ep{\discretized{\theta}_{(k+1)\stepsize}} 
    &=
    \discretized{\theta}_{k\stepsize} - \frac{\stepsize}{2 L} \sum_i
    \left( u_i\left( y_i^\rT ( \discretized{\theta}_{k \stepsize} +
    \stepsize \discretized{p}_{k \stepsize} ) \right) - u_i\left(
    y_i^\rT \discretized{\theta}_{k \stepsize} \right) \right)
    \frac{y_i}{y_i^\rT \discretized{p}_{k \stepsize}} \\ 
    &+ \left( \left( 1 + \frac{\gamma^2}{\xi^2} \right) \stepsize
    - \frac{\gamma^2}{2\xi} \stepsize^2
    - \frac{\gamma^2}{\xi^3} \left( 1 - e^{-\xi\stepsize} \right) \right)
    \discretized{p}_{k \stepsize} 
    + \left( \frac{\gamma}{\xi} \stepsize +
    \frac{\gamma}{\xi^2} \left( e^{-\xi\stepsize} - 1 \right) \right)
    \discretized{r}_{k \stepsize},
\end{align*}
\begin{align*}
   \Ep{\discretized{p}_{(k+1)\stepsize}} &= - \frac{1}{L} \sum_i
   \left( u_i\left( y_i^\rT ( \discretized{\theta}_{k \stepsize} +
   \stepsize \discretized{p}_{k \stepsize} ) \right) - u_i\left(
   y_i^\rT \discretized{\theta}_{k \stepsize} \right) \right)
   \frac{y_i}{y_i^\rT \discretized{p}_{k \stepsize}} \\ & + \left( 1 + \frac{\gamma^2}{\xi^2} \left( 1 - \xi\stepsize - e^{-\xi\stepsize} \right) \right) \discretized{p}_{k\stepsize} 
   + \frac{\gamma}{\xi} \left( 1 - e^{-\xi\stepsize} \right) \discretized{r}_{k \stepsize}.
\end{align*}
\begin{align*}
    \lefteqn{\Ep{\discretized{r}_{(k+1)\stepsize}}} \\ &= \left(
    \frac{\gamma}{L\xi} - \frac{\gamma}{L\xi^2}
    \frac{1-e^{-\xi\stepsize}}{\stepsize} \right) \cdot \sum_i \left(
    u_i\left( y_i^\rT ( \discretized{\theta}_{k \stepsize} + \stepsize
    \discretized{p}_{k \stepsize} ) \right) - u_i\left( y_i^\rT
    \discretized{\theta}_{k \stepsize} \right) \right)
    \frac{y_i}{y_i^\rT \discretized{p}_{k \stepsize}} \\ &+ \left( \frac{\gamma^3}{\xi^2}\stepsize 
    + \frac{\gamma^3}{\xi^2} \stepsize e^{-\xi\stepsize}
    - \left( \frac{2\gamma^3}{\xi^3} + \frac{\gamma}{\xi}
    \right) \left( 1 - e^{-\xi\stepsize} \right) \right) \discretized{p}_{k
      \stepsize} 
      + \left( e^{-\xi\stepsize} +\frac{\gamma^2}{\xi} \stepsize e^{-\xi\stepsize}
    - \frac{\gamma^2}{\xi^2} \left( 1 - e^{-\xi\stepsize} \right)
    \right) \discretized{r}_{k \stepsize}.
\end{align*}

Since processes $\discretized{\theta}$, $\discretized{p}$ and
$\discretized{r}$ share the same Brownian motion, we use It\^o
isometry to calculate their covariance.  For example, we can obtain
that
\begin{align*}
    &\Ep{\left(\discretized{\theta}_{(k+1)\stepsize} -
    \Ep{\discretized{\theta}_{(k+1)\stepsize}}\right)
    \left(\discretized{p}_{(k+1)\stepsize} -
    \Ep{\discretized{p}_{(k+1)\stepsize}}\right)^\rT} \\ 
    &= \frac{2\gamma^2}{\xi L} \Ep{ \left(
    \int_{k\stepsize}^{(k+1)\stepsize} \left(
    \left( (k+1)\stepsize- s \right) + \frac{1}{\xi} \left( e^{-\xi\left( (k+1)\stepsize- s \right)} - 1 \right) \right)
    \rd B_{s}^r \right)
    \left(
    \int_{k\stepsize}^{(k+1)\stepsize} 
    \left( 1 - e^{-\xi((k+1)\stepsize - s)} \right) \rd B_{s}^r 
    \right)^\rT } \\
    &= \frac{2\gamma^2}{\xi L}
    \int_{k\stepsize}^{(k+1)\stepsize} 
    \left(
    \left( (k+1)\stepsize- s \right) + \frac{1}{\xi} \left( e^{-\xi\left( (k+1)\stepsize- s \right)} - 1 \right) \right) 
    \left( 1 - e^{-\xi((k+1)\stepsize - s)} \right)
    \rd s \cdot \mI_{d \times d} \\
    &= \frac{\gamma^2}{\xi^3 L}
    \left( \xi\stepsize - \left( 1 - e^{-\xi\stepsize} \right) \right)^2 \cdot \mI_{d \times d}.
\end{align*}
Similarly, we obtain the entire covariance matrix for the tuple
$\discretized{x}_{(k+1)} = (\discretized{\theta}_{(k+1)},
\discretized{p}_{(k+1)}, \discretized{r}_{(k+1)})$:
\begin{multline}
\Ep{\left(\discretized{x}_{(k+1)\stepsize} -
  \Ep{\discretized{x}_{(k+1)\stepsize}}\right)
  \left(\discretized{x}_{(k+1)\stepsize} -
  \Ep{\discretized{x}_{(k+1)\stepsize}}\right)^\rT} \\ 
  = \frac{1}{L}
\left(\begin{array}{ccc} 
\sigma_{11} \cdot
  \mI_{d \times d} & \sigma_{12} \cdot \mI_{d
    \times d} & \sigma_{13} \cdot \mI_{d \times d} \\ \sigma_{12} \cdot \mI_{d \times d} & \sigma_{22} \cdot \mI_{d \times d} & \sigma_{23}
  \cdot \mI_{d \times d} \\ \sigma_{13} \cdot \mI_{d \times d} &
  \sigma_{23} \cdot \mI_{d \times d} & \sigma_{33} \cdot \mI_{d \times
    d}
    \end{array}\right).
\end{multline}
where
\begin{align*}
\sigma_{11} & = \frac{2\gamma^2}{\xi^3} \stepsize 
- \frac{2\gamma^2}{\xi^2} \stepsize^2
+ \frac{2\gamma^2}{3\xi} \stepsize^3
- \frac{4 \gamma^2}{\xi^3} \stepsize e^{-\stepsize  \xi }
+ \frac{\gamma^2}{\xi^4} \left( 1 - e^{-2 \xi\stepsize} \right), \\
\sigma_{12} & = \frac{\gamma^2}{\xi^3 L}
\left( \xi\stepsize - \left( 1 - e^{-\xi\stepsize} \right) \right)^2, \\
\sigma_{22} & = \frac{2 \gamma^2}{\xi} \stepsize
- \frac{4 \gamma^2}{\xi^2} \left( 1 - e^{-\xi\stepsize} \right)
+ \frac{\gamma^2}{\xi^2} \left( 1 - e^{-2\xi\stepsize} \right), \\
\sigma_{13} & = 
- \frac{ \gamma^3 }{\xi^2} \stepsize^2 \left( 2 e^{ - \xi \stepsize } + 1 \right)
+ \left(\frac{2\gamma^3}{\xi^3}
- \frac{\gamma^3}{\xi^3} e^{-2 \xi \stepsize}
- \frac{4\gamma^3}{\xi^3} e^{-\stepsize  \xi }
- \frac{ 2\gamma }{ \xi } e^{-\stepsize \xi } \right) \stepsize  
+ \left( \frac{3\gamma^3}{2 \xi^4} + \frac{\gamma}{\xi^2} \right)
\left(1 - e^{-2 \xi \stepsize}\right) , \\
\sigma_{23} & = 
\frac{\gamma^3}{\xi^2} \left(e^{-2 \xi \stepsize} - 2 e^{-\xi \stepsize} - 2\right) \stepsize 
+ \frac{3\gamma^3}{2\xi^3} \left(e^{-2\xi\stepsize} - 4e^{-\xi\stepsize} + 3\right)
+ \frac{\gamma}{\xi} \left(1 - e^{-\xi \stepsize} \right)^2, \quad \mbox{and} \\
\sigma_{33} & = 
-\frac{\gamma^4}{\xi^2} \stepsize^2 e^{-2 \xi \stepsize} 
+ \left( -\frac{2\gamma^2}{\xi} e^{-2 \xi \stepsize}
+ \frac{\gamma^4}{\xi^3} \left( - 3 e^{-2 \xi \stepsize} + 4 e^{- \xi \stepsize} + 2 \right)
\right) \stepsize \\
&+ \frac{\gamma^4}{2 \xi^4} \left(-5e^{-2 \eta  \xi } + 16 e^{-\eta  \xi }-11\right)
+ \frac{\gamma ^2}{\xi^2} \left(-3e^{-2 \eta  \xi } + 4 e^{-\eta  \xi }-1\right)
+ \left(1-e^{-2 \eta  \xi }\right).
\end{align*}


\end{document}